%% file: main.tex
\documentclass[10pt,twocolumn,letterpaper]{article}

\usepackage[ruled]{algorithm2e}

\usepackage[pagenumbers]{cvpr} 

\usepackage{booktabs}

\input{commands}

\usepackage[pagebackref,breaklinks,colorlinks]{hyperref}

\usepackage[capitalize]{cleveref}
\crefname{section}{Sec.}{Secs.}
\Crefname{section}{Section}{Sections}
\Crefname{table}{Table}{Tables}
\crefname{table}{Tab.}{Tabs.}

\def\cvprPaperID{11974} 
\def\confName{CVPR}
\def\confYear{2023}

\title{Data-Centric Debugging: mitigating model failures via targeted data collection}

\author{Sahil Singla, Atoosa Malemir Chegini, Mazda Moayeri, Soheil Feizi\\
University of Maryland\\
\small{\texttt{\{ssingla,atoocheg,mmoayeri,sfeizi\}}
\texttt{@umd.edu}}
}
\begin{document}

\twocolumn[{
    \maketitle
    \begin{center}
    \centering
    \begin{minipage}{0.15\textwidth}
    \includegraphics[trim=0cm 0.3cm 0cm 0cm, clip, width=\linewidth]{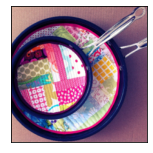}
    \vspace{-0.6cm}
    \captionof{figure}*{\textbf{label: frypan}}
    \end{minipage}\qquad 
    \begin{minipage}{0.8\textwidth}
    \includegraphics[trim=0.8cm 0.3cm 4.5cm 0cm, clip, width=\linewidth]{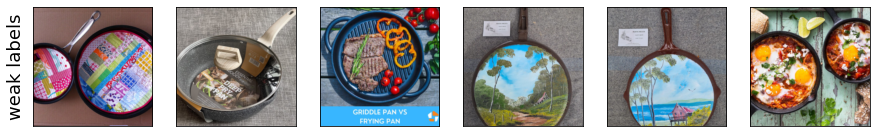}
    \vspace{-0.6cm}
    \captionof{figure}*{\textbf{visually similar images}}
    \end{minipage}\\
    \begin{minipage}{0.15\textwidth}
    \includegraphics[trim=0cm 0.3cm 0cm 0cm, clip, width=\linewidth]{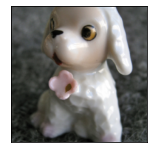}
    \vspace{-0.6cm}
    \captionof{figure}*{\textbf{label: toy poodle}}
    \end{minipage}\qquad 
    \begin{minipage}{0.8\textwidth}
    \includegraphics[trim=0.8cm 0.3cm 4.5cm 0cm, clip, width=\linewidth]{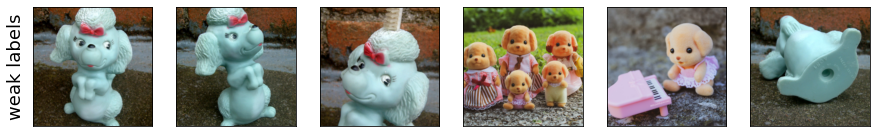}
    \vspace{-0.6cm}
    \captionof{figure}*{\textbf{visually similar images}}
    \end{minipage}
    \vspace{0.0cm}
    \captionof{figure}{We introduce a method for debugging model failures by discovering ``visually similar'' images from the web. Left: sample images on which 20 ImageNet trained models with near state-of-the-art accuracy make incorrect predictions (label below each image). Our proposed debugging methodology can fix such failure modes. Right: ``visually similar'' images from the pool images $\flickr$.}
    \label{fig:teaser}
    \end{center}
}]

\begin{abstract}
Deep neural networks can be unreliable in the real world when the training set does not adequately cover all the settings where they are deployed. Focusing on image classification, we consider the setting where we have an error distribution $\dist$ representing a deployment scenario where the model fails. We have access to a small set of samples $\dist_{sample}$ from $\dist$ and it can be expensive to obtain additional samples. In the traditional model development framework, mitigating failures of the model in $\dist$ can be challenging and is often done in an ad hoc manner. In this paper, we propose a general methodology for model debugging that can systemically improve model performance on $\dist$ while maintaining its performance on the original test set. Our key assumption is that we have access to a large pool of weakly (noisily) labeled data $\flickr$. However, naively adding $\flickr$ to the training would hurt model performance due to the large extent of label noise. Our Data-Centric Debugging (DCD) framework carefully creates a debug-train set by selecting images from $\flickr$ that are perceptually similar to the images in $\dist_{sample}$. To do this, we use the $\ell_2$ distance in the feature space (penultimate layer activations) of various models including ResNet, Robust ResNet and DINO where we observe DINO ViTs are significantly better at discovering similar images compared to Resnets. Compared to LPIPS, we find that our method reduces compute and storage requirements by 99.58\%. Compared to the baselines that maintain model performance on the test set, we achieve significantly (+9.45\%) improved results on the debug-heldout sets.
\end{abstract}

\section{Introduction}\label{sec:intro}

\begin{figure*}[ht!]
        \centering
        \includegraphics[trim=0cm 0cm 0cm 0cm, clip, width=\linewidth]{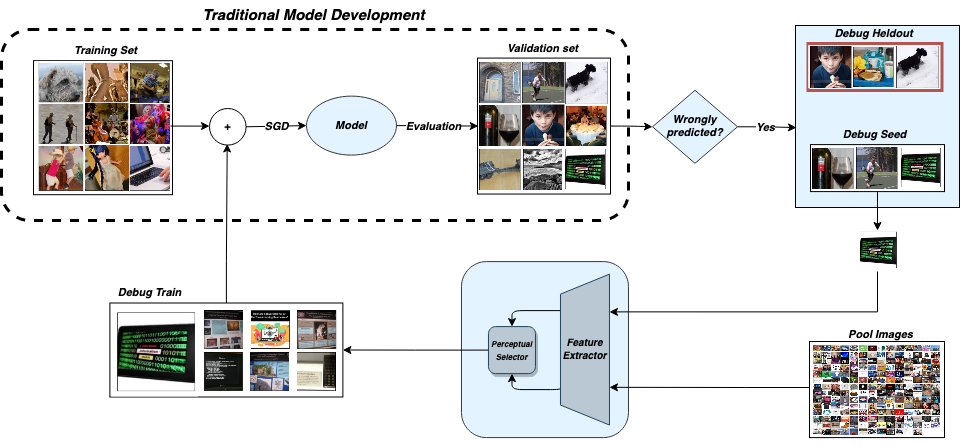}
        \label{fig:framework}
    \caption{Our framework (Data-Centric Debugging) for model debugging via targeted image retrieval. We want to improve model performance on an error distribution $\dist$ (while maintaining accuracy on the test sets) using a sample $\dist_{sample}$ (from $\dist$) and a weakly labeled dataset $\flickr$. We divide the set $\dist_{sample}$ into two disjoint sets: $\dist_{seed}$ and $\dist_{heldout}$. Using the samples in $\dist_{seed}$, we want to add several ``visually similar'' images (selected from $\flickr$) to the training set, such that the model performance improves on $\dist_{heldout}$.}
    \label{fig:main_fig}
\end{figure*}


As machine learning systems are increasingly being deployed in the real world, understanding and mitigating their failure modes becomes critical to ensure that models work reliably in different deployment settings. For example, in medical applications, it is common to train a model using data from a few hospitals, and then deploy it more broadly to hospitals outside the training set\cite{Zech2018VariableGP}. In such cases, we may want to identify the hospital systems on which the model fails and feed more training data from those systems into the model to improve its performance. 

Most of the prior works in the literature focus on mitigating a small set of failure modes identified by a human-in-the-loop \cite{santurkar2021editing, mitchell2021fast}.  This can involve collecting new datasets with objects in uncommon settings \cite{hendrycks2018benchmarking, objectnet19, hendrycks2021nae, 9710159, kattakinda22a} (e.g. frog in snow, ship indoors) which can be time consuming and expensive. Moreover, in several cases, humans might not be adequately aware of the undesirable failure modes and even if they are, collecting large number of images in the desired deployment scenarios might not always be feasible. 

In this work, we focus on the image classification problem $\cX \to \cY$ where the goal is to predict the ground truth label $y \in \cY$ for input $\bx \in \cX$. In the traditional model development framework, we have a training set, a validation set and a test set. The training set is used to train the model, the validation set is used to evaluate and improve the model performance during development and the test set is used to report a final metric for the model performance. 

In this setting, we consider an error distribution $\dist$ representing a deployment scenario where a trained model fails i.e. the model makes incorrect predictions on every sample from $\dist$. We have access to a small set of images $\dist_{sample}$ from $\dist$ and it is prohibitively expensive to obtain more samples from $\dist$. Our goal is to improve model performance on $\dist$ while maintaining performance on the existing test set(s). One naive approach could be to add $\dist_{sample}$ to the training set. However, the model might overfit to $\dist_{sample}$ and fail to generalize to novel samples from $\dist$. Thus, the traditional development framework can be ineffective when sampling a large amount of data from $\dist$ is infeasible. 

\begin{table*}[t]
    \centering
    \renewcommand{\arraystretch}{1.3} 
    \centering
    \begin{tabular}{ l | ll | lll |  lll }
    \toprule
     \multirow{2}{1.5cm}{\textbf{Debug-Train}} & \multicolumn{8}{|c}{\textbf{Accuracy on different sets}} \\ 
    \cmidrule{2-9}
       & \multicolumn{2}{|c|}{\textbf{incorrectly classified}} 
      & \multicolumn{3}{c|}{\textbf{subset of 160 classes}} & \multicolumn{3}{c}{\textbf{all 1000 classes}} \\ 
     & Seed & Heldout & MFreq & Compl. & INet &  MFreq & Compl. & INet \\
\midrule
original & 0\% & 0\% & 35.56\% & 56.89\% & 62.89\% &  63.70\% & 76.12\% & 76.47\% \\
DCD-Complete & 18.53\% & 23.09\% & 37.56\% & 53.84\% & 49.22\% & 61.70\% & 72.93\% & 75.07\%  \\
DCD-Random & 16.78\% & 20.17\% & 41.56\% & 58.81\% & 63.91\% & 63.77\% & 75.21\% & 76.37\%  \\
\midrule
DCD-DINO & 36.28\% & \textbf{29.62\%} & 54.06\% & \textbf{63.85\%} & \textbf{64.62\%} & 65.28\% & \textbf{76.42\%} & 76.54\% \\
\bottomrule
\end{tabular}
\caption{Results using DINO ViT-S/8 in our framework. ``INet'' denotes the ImageNet test set, ``MFreq'' denotes the ImageNet-V2\cite{pmlr-v97-recht19a} MatchedFrequency set, ``Compl.'' (Complement) denotes the set of all ImageNet-V2 images excluding ``MFreq'' images. }
\label{table:intro_summary}
\end{table*}

In this work, we propose a new formulation where in addition to the training data and $\dist_{sample}$, we have access to a large weakly-labeled (i.e., very noisily labeled) pool of images denoted by $\flickr$. Here, $\flickr$ could be obtained from Flickr, Commoncrawl \cite{commoncrawl} or any suitable data source. We collect $\flickr$ using Flickr search (Section \ref{sec:web_data}) and carry out a filtration step to ensure that the images in $\flickr$ are ``significantly different'' from the test set (Section \ref{sec:remove_test_set}). Because of the noisy labeling, we find that naively adding the complete set $\flickr$ to the training set can hurt model performance. For example, we observe $-13.67\%$ drop in accuracy (Table \ref{table:single_model}). Thus, we want to select a few samples from $\flickr$ \textit{without human supervision} to improve model performance on $\dist$. 

Intuitively, by selecting several images from $\flickr$ that are ``visually similar'' to the images in $\dist_{sample}$, we would expect a broader coverage of $\dist$ resulting in improved model performance compared to say, only adding $\dist_{sample}$. However, because $\flickr$ can be large, identifying such similar images can be difficult. Previous works on similarity matching such as LPIPS \cite{zhang2018unreasonable} compute very high dimensional image embeddings and use the $l_{2}$ distance between them as the ``perceptual similarity distance''. However, their high dimensionality makes them infeasible for large-scale visual search problems (discussed in Section \ref{sec:visual_distances}).

Moreover, even if we discover the similar images, we may achieve improved results on $\dist_{sample}$ due to some patterns specific to the $\dist_{sample}$ images. For example, an image in $\dist_{sample}$ contains some pattern that is similar to the patterns of some different class and similarity matching may discover images from the other class. As a result, the model may achieve improvements on $\dist_{sample}$ and still fail to generalize to new samples from $\dist$. Thus, to ensure that a model revision improves performance over the distribution $\dist$ as opposed to simply the observed instances $\dist_{sample}$, a careful framework for model debugging is required.

To address these challenges, we introduce {\bf Data-Centric Debugging (DCD)}, illustrated in Figure \ref{fig:main_fig}: a framework for targeted data collection to mitigate failure modes of deep models and faithfully assess model performance on the error distribution. To retrieve visually similar images, we use the $\ell_2$ distance in the feature space (penultimate layer activations) of a deep network. Compared to LPIPS \cite{zhang2018unreasonable} (state of the art for measuring perceptually similarity), our embeddings have several orders of magnitude lower dimensionality and require vastly less memory and processing time to retrieve similar images. We find that our method reduces compute and storage requirements by \textbf{99.58\%} (Table \ref{table:lpips_ours_experiments}).   

To faithfully evaluate model performance, we divide the set $\dist_{sample}$ into two disjoint sets namely, $\dist_{seed}$ and $\dist_{heldout}$. We refer to them as debug-seed set and debug-heldout sets respectively. We want to use the set $\dist_{seed}$ for discovering visually similar images such that the model performance improves on $\dist_{heldout}$. That is, we only use $\dist_{seed}$ (not $\dist_{heldout}$) for visual similarity matching and evaluate model performance on $\dist_{heldout}$. Because $\dist_{heldout}$ is disjoint from $\dist_{seed}$ but from the same distribution $\dist$, an improved performance would suggest that the model is not overfitting to the images in $\dist_{seed}$ and can generalize to novel samples from $\dist$. Thus, model performance on $\dist_{heldout}$ is a more faithful evaluation metric for $\dist$. 

We apply our proposed framework on the ImageNet \cite{5206848} classification task. We first select 160 ImageNet classes on which 20 highly accurate ImageNet trained models achieve low accuracy (details in Appendix \ref{sec:appendix_160_classes}). From these classes, we select the incorrectly classified samples from the ImageNet-V2 dataset as the $\dist_{sample}$ set. Next, we divide $\dist_{sample}$ into the $\dist_{seed}$ and $\dist_{heldout}$ sets (Section \ref{sec:seed_heldout_sets}). For an image $\bx \in \dist_{seed}$ with label $i$, we can either select visually similar images from the subset of $\flickr$ with weak label $i$ (denoted by $\flickr(i)$) or from the complete set $\flickr$ thereby discarding the weak labels. In the latter case, we can assign the label $i$ to selected images. We find that selecting from the complete set often leads to images that are similar to $\bx$, but from a different class, thereby contaminating the dataset with wrongly labeled images. We illustrate this in Figure \ref{fig:main_comparison}. Thus, we select similar images from the subset $\flickr(i)$.  

In Section \ref{sec:experiments}, we experiment with several different models for extracting image embeddings for visual similarity matching namely, Standard Resnet-50, Robust Resnet-50, DINO ViT-S/8 and DINO ViT-S/16 \cite{dino}. Our experiments (Table \ref{table:single_model}) suggest that DINO models are significantly better at discovering similar images compared to Resnets. 

In Table \ref{table:intro_summary}, we compare our method against the ``original'' model trained using standard ImageNet, ``DCD-random'': trained using randomly selected images from subsets $\flickr(i)$, ``DCD-complete'': trained on the full $\flickr$ dataset (with class re-weighting so that weights assigned to classes are same as in ImageNet). For our results (``DCD-DINO''), we used DINO ViT S/8 for similarity matching. We observe that our method achieves the best results on Heldout set: (29.62\%), significantly outperforming both complete (23.07\%) and random (20.17\%). We also achieve the best results on several 160 class ImageNet subsets. Moreover, ``DCD-complete'' results in large accuracy drop on INet (160 classes) from 62.89\% to 49.22\% (-13.67\%) whereas with our method, the accuracy improves to 64.62\% (+1.73\%). These results highlight that our proposed framework is effective in mitigating model failures. 





In summary, we make the following contributions:
   \begin{enumerate}[leftmargin=*]
     \item We proposed {\bf DCD}, a framework for mitigating model failures via data-centric debugging. In contrast to the traditional model development using training/validation/test splits, we construct debug seed/train/heldout datasets to systematically improve failure modes of the model.
     
     \item We use the $\ell_2$ distance in the feature space for efficiently retrieving perceptually similar images to a reference image from the large pool dataset $\flickr$. Our experiments suggest that DINO models are significantly better at discovering similar images compared to Resnets. 
     
     \item Using our framework, we achieve 29.62\% accuracy on the debug-heldout set, compared to 0\%, 23.09\%, 20.17\% for the baseline models. We also achieve significant improvements: 63.85\% vs 58.81\% for runner-up (+5.04\%) on the ImageNet-V2 subset (``Complement'' in Table \ref{table:intro_summary}). 
     
   \end{enumerate}

\begin{table}
  \centering
  \begin{tabular}{lll}
    \toprule
     & LPIPS & \textbf{Ours} \\
    \midrule
    Time (secs) & 1147.58 & \textbf{4.81} \\
    Space (GBs) & 1806.74 & \textbf{7.63} \\
    \# of dims & 484992 & \textbf{2048} \\
    \bottomrule
  \end{tabular}
  \caption{Comparison between LPIPS and our method for finding similar images from a set of $1$ million images. Using our method, leads to a \textbf{99.58\% reduction in both time and space.}}
  \label{table:lpips_ours_experiments}
\end{table}

\section{Notation}
Let set $\set$ consist of (image, label) pairs: $(\bx, y) \in \set$, and $\set(i)$ denote the images with label $i$: 
$$\set(i) = \{\bx: (\bx, i) \in \set \}$$
Given two sets: $\set$ and $\mathcal{B}$, we use: $(\set - \mathcal{B})(i)$ to denote $\set(i) - \mathcal{B}(i)$.
We use $|\set|$ to denote the cardinality (number of elements) of $\set$, $[n]$ to denote the set: $[0, 1, \dotsc, n-1]$ and $\|\bz\|$ for the $l_{2}$ norm of vector $\bz$. For image $\bx$, $\Phi(\bx)$ denotes the penultimate layer output of the model $\Phi$. We use $\allclasses$ to denote the set of all 1000 ImageNet classes.

\section{Framework for model debugging}\label{sec:framework}

Consider the image classification problem $\cX \to \cY$ where we want to predict the ground truth label $y \in \cY$ for input $\bx \in \cX$. Given a trained model, we have an \textit{error distribution} $\dist$ of incorrectly classified images i.e. every image sampled from $\dist$ is misclassified by the model. We have access to a set of samples $\dist_{sample}$ from $\dist$ and it is very expensive to draw more samples. Here, $\dist$ represents the deployment scenario where we we want to improve model performance. For example, we may be interested in images with people of color, specific gender or distribution shift (e.g., people wearing masks during COVID-19), etc. 

We also assume that we have access to a large pool of weakly labeled images (noisy labels) denoted by $\flickr$. Here, $\flickr$ can be obtained using any suitable data source depending on the problem. Using $\flickr$, we want to improve model performance on $\dist$ while maintaining on the desired test set(s). 

One naive method could be to add the complete set $\flickr$ to the training set. However, because the labels in $\flickr$ can be very noisy, this can reduce the quality of the dataset and hurt model performance.  Thus, we want to improve performance on $\dist$ by selecting new training images from $\flickr$ while maintaining model performance on the desired test set(s). 

\begin{figure}
    \centering
    \includegraphics[width=\linewidth]{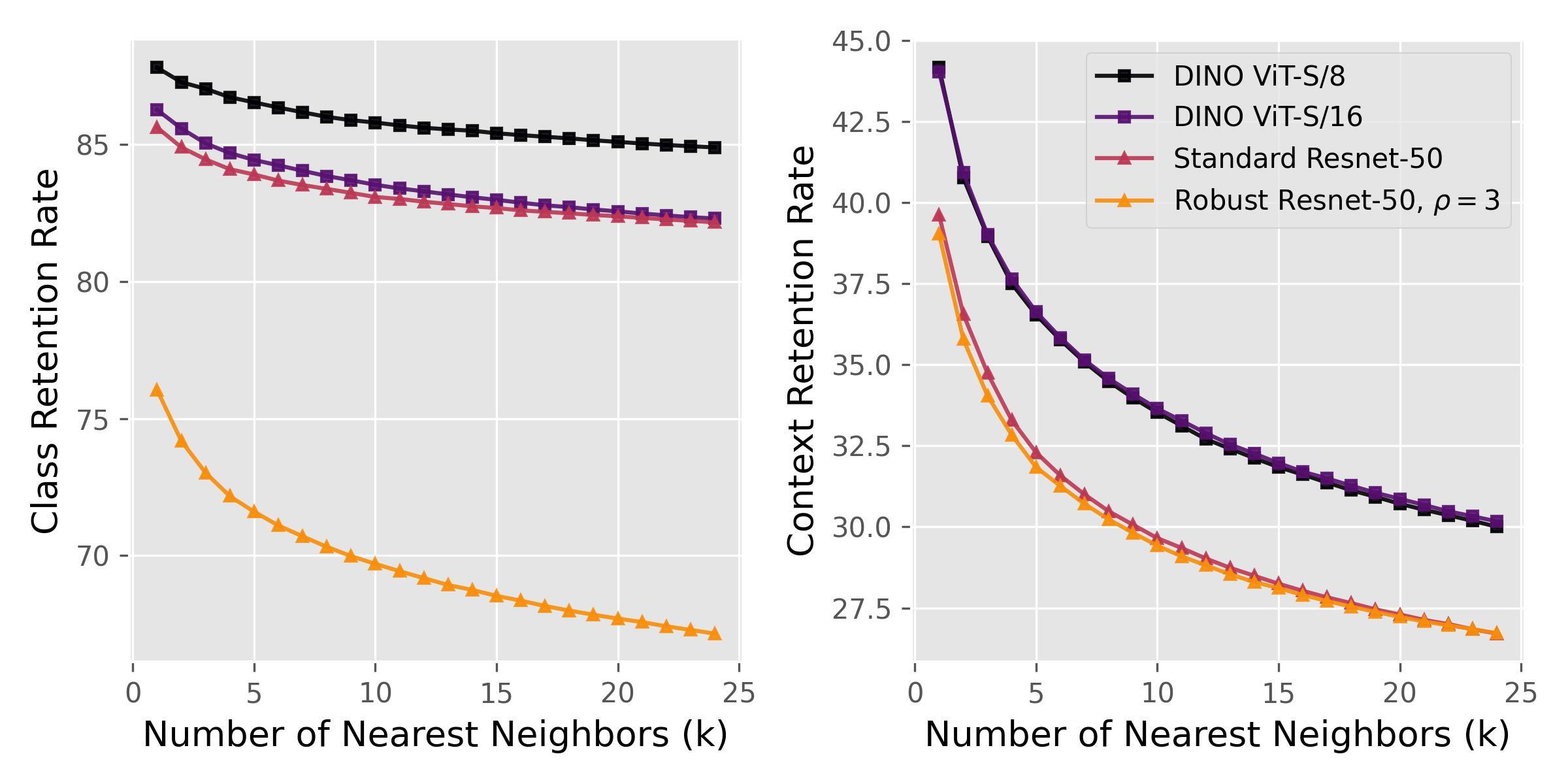}
    \caption{Class and context retention for nearest-neigbhor retrieval on the FOCUS dataset, using different models $\Phi$. DINO ViTs are superior for both class andd context retention.}
    \label{fig:focus}
\end{figure}

\begin{figure}
    \begin{subfigure}{0.22\linewidth}
    \centering
    \includegraphics[width=\linewidth]{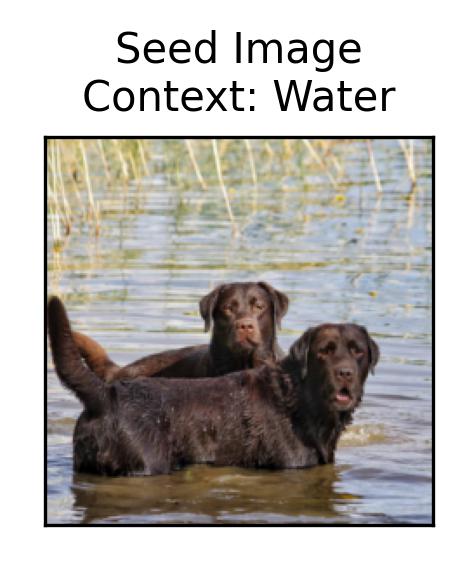}
    \vspace{0.45in}
    \end{subfigure}
    \begin{subfigure}{0.75\linewidth}
        \centering
        \includegraphics[width=\linewidth]{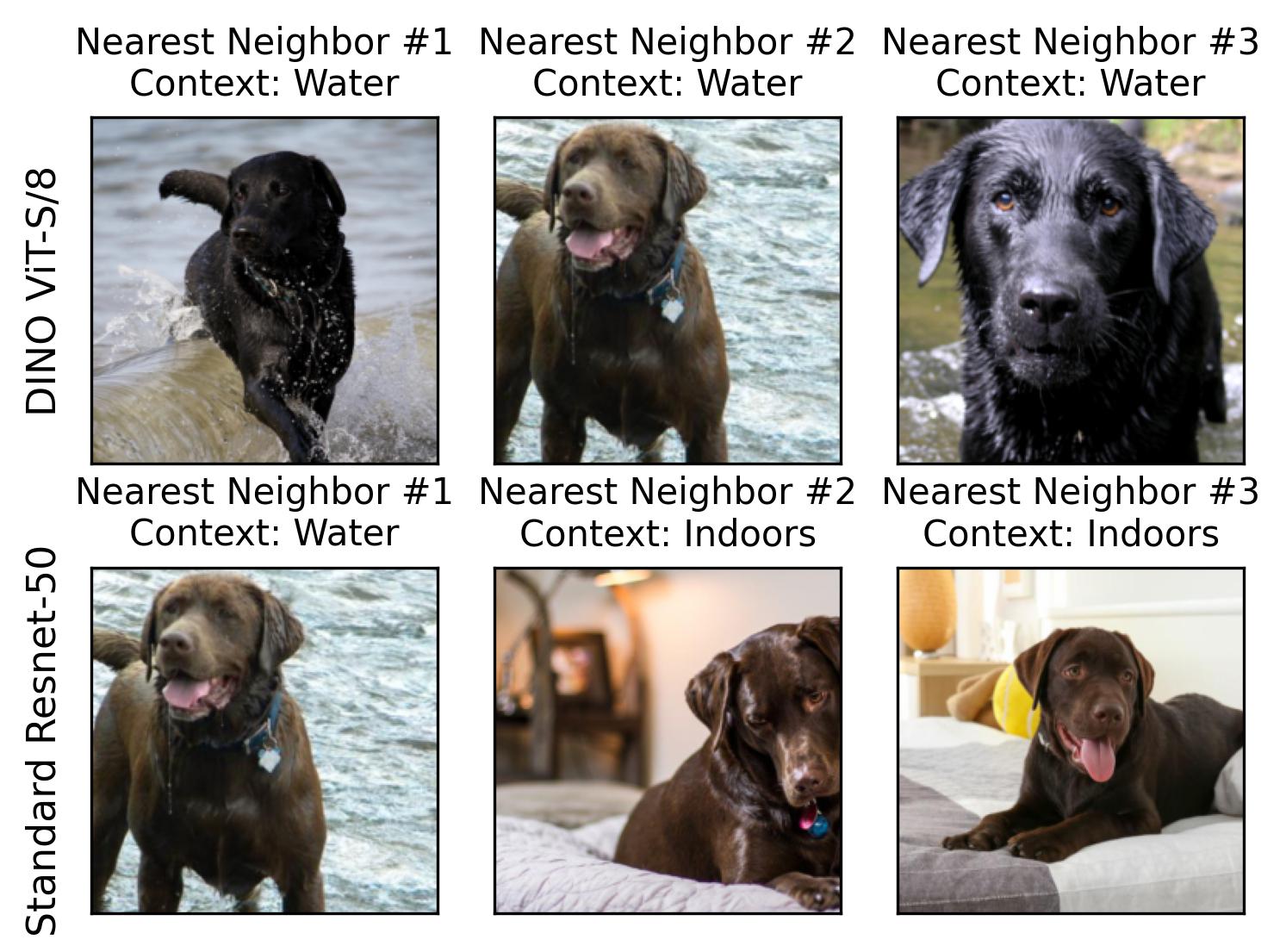}
    \end{subfigure}
    \caption{Example where nearest neighbor ($k=3$) search on FOCUS dataset using DINO ViT-S/8 features retains context, while Standard Resnet-50 only does so in one of three cases.}
    \label{fig:focus_eg}
\end{figure}

\begin{figure*}
        \centering
        \captionsetup{type=figure}
        \begin{minipage}{0.16\textwidth}
        \hspace*{0.4cm}
        \begin{subfigure}{1\linewidth}
        \includegraphics[trim=0cm 0cm 0cm 0cm, clip, width=\linewidth]{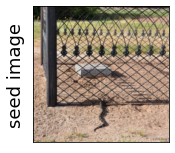}
        \caption*{\quad\ \  \textbf{label: king snake}}
        \end{subfigure}
        \end{minipage}
        \hfill
        \begin{minipage}{0.80\textwidth}
        \begin{subfigure}{\linewidth}
        \includegraphics[trim=0cm 0cm 0cm 0cm, clip, width=\linewidth]{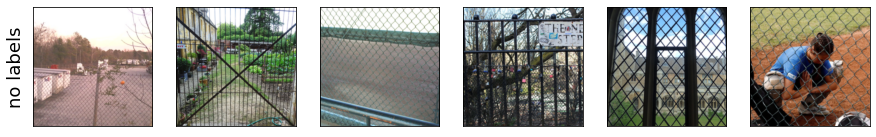}
        \end{subfigure}
        \begin{subfigure}{\linewidth}
        \vspace{0.1cm}
        \includegraphics[trim=0cm 0cm 0cm 0cm, clip, width=\linewidth]{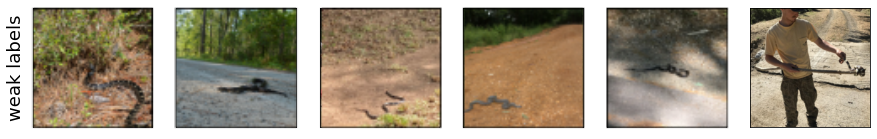}
        \end{subfigure}
        \end{minipage}
        \vspace{-0.1cm}
        \caption{left: wrongly predicted image with label \textbf{king snake}, right: similar images obtained using DINO ViT-S/8 model with two different methods. Selecting images without using any weak labels outputs samples where the main object, in this case, king snake, is absent, which can introduce wrongly labeled images in the dataset. More examples are in Appendix \ref{sec:appendix_compare_weaklabel_nolabel}.}
    \label{fig:main_comparison}
\end{figure*}

Intuitively, we would expect an improved performance on $\dist$ by selecting several images from $\flickr$ that are ``visually similar'' to $\dist_{sample}$ images. However, identifying similar images from a large dataset can be difficult. Moreover, even if we successfully discover such images, we may see an improved performance on $\dist_{sample}$ due to some patterns specific to $\dist_{sample}$ images that may fail to generalize to new images from $\dist$. For example,  the pattern in one image may be a strong match for another image from different class (see Figure \ref{fig:main_comparison}). Thus, we want an evaluation procedure that reflects the true model performance on $\dist$ because the performance on $\dist_{sample}$ may be an overestimate of the same. 

To address these challenges, we introduce {\bf Data-Centric Debugging (DCD)}, illustrated in Figure \ref{fig:main_fig}: a framework for targeted data collection to mitigate failure modes of deep models and faithfully assess model performance on the error distribution. In the next subsections, we discuss the building blocks of our framework. 

\subsection{Visual similarity matching at scale}\label{sec:visual_distances}
From the set $\flickr$, we would like to identify images that are ``visually similar'' to the images in $\dist_{sample}$ and add to the training set as we would expect such images to be most effective for improving model performance. However, finding visually similar images to a reference image from a large pool of images can be difficult. Previous work on similarity matching, LPIPS \cite{zhang2018unreasonable} uses image embeddings that are very high dimensional. For example, for an image of size $224 \times 224 \times 3$ ($150528$ dimensions), the LPIPS embedding using a pretrained AlexNet \cite{Krizhevsky2012ImageNetCW} model is of size $484,992$ ($\approx 3.22$ times the size of the original image). This leads to significant challenges because storing the LPIPS embedding takes a large amount of space and computing the LPIPS distance of a large number of images from some reference image can be very time-consuming.

In this work, we propose to use the penultimate layer output of a deep model $\Phi$ as the image embedding which is very compact (2048 for Resnet-50). Given two images $\bx, \bz$, we use the squared $l_2$ distance $(\|\Phi(\bx) - \Phi(\bz)\|^{2})$ in this space as the visual similarity distance to discover visually similar images. 
In Table \ref{table:lpips_ours_experiments}, we show that compared to LPIPS, our embedding leads to $99.58\%$ reduction in both time and storage space, thereby making it feasible to obtain  visually similar images from a large pool of images. 

We experiment with four pretrained models $\Phi$ for computing these distances: Standard Resnet-50, Robust Resnet-50, DINO ViT-S/16 and DINO ViT-S/8 \cite{dino} (Appendix \ref{sec:appendix_similarity_matching}). To better inform the choice of the model $\Phi$, we conduct an experiment on the FOCUS dataset \cite{kattakinda22a}. FOCUS consists of common objects in various settings, leading to {\it two} labels per  sample: one label denoting class (bird, plane, etc), and the other denoting context (snow, night, indoors, etc). We obtain features for every sample in FOCUS using various backbones, and then obtain $k=25$ nearest neighbors per sample in the feature space of each model. We then compute the percent of neighbors amongst the top $k'\leq k$ that retain (i) object class and (ii) image context.

Figure \ref{fig:focus} visualizes the results. We observe that the retention rate for class is far higher than for context. Retaining both class and context is important for our use case as the set $\dist_{sample}$ may consist of instances of a class in an uncommon context and we would want to select more examples of that object in that same context. We find that DINO vision transformers are superior in {\it both} class and context retention compared to Resnet models. Specifically, DINO ViT-S/8 achieves the best class and context retention. This unique property makes DINO transformers a prime candidate for visual similarity computation and targeted image retrieval. In Figure \ref{fig:focus_eg}, we show an example where the 3 nearest neighbors using DINO ViT-S/8 features retain context, but Standard Resnet-50 features do not. 


\subsection{Collecting large pool of images from the web}\label{sec:web_data}

We want to collect a large pool of images from the web and identify images that are visually similar to the images in $\dist_{sample}$. Since collecting a large number of images for all 1000 ImageNet classes can be time consuming, we first select 160 classes (denoted by $\errorclasses$) on which $20$ highly accurate ImageNet trained models achieve low accuracy (see Appendix \ref{sec:appendix_160_classes}). For each class $i \in \errorclasses$, we obtain their synset (set of synonyms). For example, in the synset \{`junco', `snowbird'\}, `junco' and `snowbird' are synonyms. For each synonym in the synset, we perform a Flickr search and collect the URLs of the first 30,000 images in the search results. After collecting URLs for all classes in $\errorclasses$, we remove URLs that were common across multiple classes. This results in a weakly labeled dataset (denoted by $\bar{\flickr}$) consisting of 952,951 images across 160 classes. Note that $\bar{\flickr}$ is weakly labeled because all images in the search results may not contain the relevant object in the search term. 






\begin{table}[t]
\centering
\begin{tabular}{{l | l | l }}
\toprule
\textbf{Dataset} & \textbf{Class subset} & \textbf{\# of images} \\
\midrule
Seed & \multirow{2}{*}{160 classes ($\errorclasses$)} & 1,031 \\
Heldout &  & 719 \\
\midrule
\multirow{2}{*}{MFreq} & 160 classes ($\errorclasses$) & 1,600 \\
 & 1000 classes ($\allclasses$) & 10,000 \\
 \midrule
\multirow{2}{1.2cm}{Comple-ment} & 160 classes ($\errorclasses$) & 1,668 \\
 & 1000 classes ($\allclasses$) & 10,683 \\
\bottomrule
\end{tabular}
\caption{Dataset sizes. Seed and Heldout are for single-model.}
\label{table:statistics_single_model}
\end{table}

\subsection{Removing images visually similar to test sets}\label{sec:remove_test_set}
Since the model performance on test set can be trivially improved by adding images from the test set to the training set, it is critically important to ensure that the new images added to the training set are ``sufficiently different'' from the test set images. To this end, we introduce a filtration step based on the criteria that the newly added images should be at least as different from test set images as they are between the ImageNet train/test sets (see Appendix \ref{sec:appendix_remove_test_set}). Thus for each class $i \in \errorclasses$, we first compute a threshold visual similarity distance $\tau(i)$ using the ImageNet dataset. 

Let $\union$ denote the union of all test sets 
that we want to evaluate our model on. This includes the seed set, heldout set, and all test sets. We select the images $\bx \in \bar{\flickr}(i)$ that have visual similarity distance $> \tau(i)$,\ from all images $\bz \in \union(i)$. The new dataset constructed by selecting such images is denoted by $\flickr$. We select ``visually similar'' from $\flickr$ to prevent images identical to the test set from being selected.

\subsection{Debug-seed and -heldout sets}\label{sec:seed_heldout_sets}

We divide the set $\dist_{sample}$ into two disjoint sets: $\dist_{seed}$ and $\dist_{heldout}$. Using $\dist_{seed}$, we want to add images from $\flickr$ to the training set that result in improved model performance on $\dist_{heldout}$. We stress that $\dist_{heldout}$ is never used for data collection. The intuition here is that since we are only using $\dist_{seed}$ to collect new images and $\dist_{heldout}$ is from the same distribution $\dist$, an improved performance would suggest that the model is learning relevant concepts (not overfitting to $\dist_{seed}$). This leads to the following definitions: 

\begin{itemize}[leftmargin=*]
    \item \textbf{Debug-Seed Set}: set of images ($\dist_{seed}$) sampled from $\dist$ used to collect new training data to improve performance
    \item \textbf{Debug-Heldout Set}: set of images ($\dist_{heldout}$) sampled from $\dist$ disjoint from $\dist_{seed}$ that is used to evaluate performance of model trained on images collected using $\dist_{seed}$. 
\end{itemize}



We remark that this is similar to the validation/test setup in model development. 
We construct the debug-seed and heldout sets in two settings, (a) single-model: images incorrectly classified by a single model (We use a Standard Resnet-50) and (b) multiple-model: images incorrectly classified by 20 highly accurate models (see Appendix \ref{sec:appendix_all_model}). In both these settings, we use the 160 class subset ($\errorclasses$) for which we obtained Flickr images (Section \ref{sec:web_data}).

%


We use the ImageNet-V2 dataset \cite{pmlr-v97-recht19a} to sample the seed $\dist_{seed}$ and heldout $\dist_{heldout}$ sets. ImageNet-V2 consists of three (non-disjoint) sets namely, (a) MatchedFrequency, (b) Threshold0.7, (c) TopImages. We observe that models achieve the lowest accuracy on MatchedFrequency (or ``MFreq'') set \cite{pmlr-v97-recht19a}. Thus, we use the incorrectly classified images from this set to construct the set $\dist_{seed}$. This ensures that $\dist_{seed}$ has a large size. We want the heldout set $\dist_{heldout}$ to be disjoint from $\dist_{seed}$. Thus, we take the union of all the three sets and remove the ``MFreq'' images from the union to define the ``Complement'' set. 


In the single-model setting, we construct the seed set ($\dist_{seed}$) by selecting images from ``MFreq'' with labels in $\errorclasses$ that were incorrectly classified by the Resnet-$50$ model. For the heldout set ($\dist_{heldout}$), we select the incorrectly classified images from the ``Complement'' set, again from the 160 classes in $\errorclasses$. The sizes of these datasets are in Table \ref{table:statistics_single_model}.

In the multiple-model setting, the procedure is similar except that we select images from $160$ classes that were incorrectly classified by each the 20 models. 





\subsection{Debug-train and -validation sets}\label{sec:debug_validate_train_construct}


We use the images in $\dist_{seed}$ to select new images from $\flickr$ and add them to the training set. We may also want to validate that upon training the model on these new images, the performance improves on images visually similar to $\dist_{seed}$. Thus, we define the debug-train and debug-validation sets:
\begin{itemize}[leftmargin=*]
    \item \textbf{Debug-Train Set}: set of images selected from $\flickr$ and added to the training set to improve model performance.
    \item \textbf{Debug-Validation (De-Val) Set}: set of images selected from $\flickr$ (and disjoint from debug-train set) to validate that the model performance improves on images visually similar to the images in $\dist_{seed}$.
\end{itemize}


\begin{table*}[t]
    \centering
    \renewcommand{\arraystretch}{1.3} 
    \centering
    \begin{tabular}{l | ll | lll |  lll }
    \toprule
    \multirow{3}{3cm}{\textbf{Debug-Train method}} & \multicolumn{8}{|c}{\textbf{Accuracy on different sets}} \\ 
    \cmidrule{2-9}
    & \multicolumn{2}{|c|}{\textbf{incorrectly classified}} 
      & \multicolumn{3}{c|}{\textbf{subset of 160 classes}} & \multicolumn{3}{c}{\textbf{all 1000 classes}} \\ 
     & Seed & Heldout & MFreq & Compl. & INet &  MFreq & Compl. & INet \\
\midrule
original & 0\% & 0\% & 35.56\% & 56.89\% & 62.89\% &  63.70\% & 76.12\% & 76.47\% \\
DCD-Complete & 18.53\% & 23.09\% & 37.56\% & 53.84\% & 49.22\% & 61.70\% & 72.93\% & 75.07\%  \\
DCD-Random & 16.78\% & 20.17\% & 41.56\% & 58.81\% & 63.91\% & 63.77\% & 75.21\% & 76.37\%  \\
\midrule
DCD-DINO (ViT-S/8) & \multirow{1}{*}{36.28\%} & \multirow{1}{*}{\textbf{29.62\%}} & \multirow{1}{*}{54.06\%} & \multirow{1}{*}{\textbf{63.85\%}} & \multirow{1}{*}{\textbf{64.62\%}} & \multirow{1}{*}{65.28\%} & \multirow{1}{*}{\textbf{76.42\%}} & \multirow{1}{*}{76.54\%} \\
{DCD-DINO (ViT-S/16)}  & \multirow{1}{*}{\textbf{36.76\%}} & \multirow{1}{*}{26.98\%} & \multirow{1}{*}{\textbf{55.00\%}} & \multirow{1}{*}{62.83\%} & \multirow{1}{*}{64.11\%} & \multirow{1}{*}{\textbf{65.69\%}} & \multirow{1}{*}{75.38\%} & \multirow{1}{*}{76.41\%} \\
{DCD-Resnet (Standard)}  & \multirow{1}{*}{32.39\%} & \multirow{1}{*}{28.09\%} & \multirow{1}{*}{51.75\%} & \multirow{1}{*}{63.31\%} & \multirow{1}{*}{64.57\%} & \multirow{1}{*}{65.62\%} & \multirow{1}{*}{75.96\%} & \multirow{1}{*}{\textbf{76.70\%}} \\
{DCD-Resnet (Robust)} & \multirow{1}{*}{33.07\%} & \multirow{1}{*}{26.84\%} & \multirow{1}{*}{51.5\%} & \multirow{1}{*}{62.29\%} & \multirow{1}{*}{62.27\%} & \multirow{1}{*}{65.04\%} & \multirow{1}{*}{75.70\%} & \multirow{1}{*}{76.50\%}\\
\bottomrule
\end{tabular}
\caption{Results for \emph{single-model} seed and heldout sets. ``INet'' denotes the ImageNet test set, ``MFreq'' denotes the ImageNet-V2\cite{pmlr-v97-recht19a} MatchedFrequency set, ``Compl.'' denotes the Complement set i.e. all ImageNet-V2 images excluding ``MFreq'' images. }
\label{table:single_model}
\end{table*}

For each image $\bx \in \dist_{seed}(i)$, the de-val should contain a set of few (say $k$) images visually similar to $\bx$ with label $i$. We may construct this set by selecting $k$ images with the smallest visual similarity distance to $\bx$ from $\flickr(i)$. However, for two images $\bx, \bz \in \dist_{seed}(i)$, the sets of $k$ images may overlap. Thus, some seed images may have fewer (than $k$) images included and not be well represented. To address this limitation, we use an algorithm (in Appendix \ref{sec:appendix_sec_algo_non_disjoint}) that removes the selected images on the fly and avoids overlaps. The resulting de-val set is denoted by $\val$. We construct the debug-train set using a similar procedure. Because we want debug-train set to be disjoint from the debug-val set ($\val$), we find visually similar images from the subset: $(\flickr - \val)(i)$. The procedure is same as the de-val set except that we use: $(\flickr - \val)$ instead of $\flickr$. We first construct the de-val set using $k=4$ followed by the debug-train set using $k=46$. 

\section{Experiments}\label{sec:experiments}

In this Section, we discuss results using the single-model seed and heldout sets (Section \ref{sec:seed_heldout_sets}). Results for multiple-model are similar and discussed in Appendix \ref{sec:appendix_multiple_model}. We evaluate our proposed method on two criteria: the improvement in accuracy on the heldout-debug set and the accuracy drop on the ImageNet, MFreq and Complement test sets. For each of these test sets, we evaluate on both 160 class subset and all 1000 classes. We use the Resnet-50 architecture for training all models. Each model was trained for 90 epochs over eight GPUs (RTX 2080 Ti). We use the composer library for training all models to reduce training time \cite{mosaicml2022composer}.


\begin{figure*}
        \centering
        \captionsetup{type=figure}
        \begin{subfigure}{0.3\linewidth}
        \includegraphics[trim=0cm 0cm 0cm 0cm, clip, width=\linewidth]{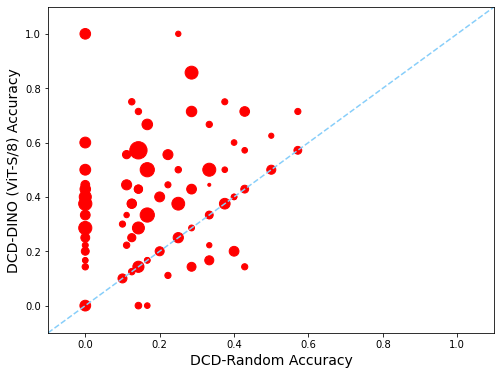}
        \vspace{-0.05cm}
        \caption{accuracy for 160 classes for DCD-DINO (Y-axis) and -Random (X-axis) on seed set}
        \label{fig:seed_160}
        \end{subfigure}\qquad
        \begin{subfigure}{0.3\linewidth}
        \includegraphics[trim=0cm 0cm 0cm 0cm, clip, width=\linewidth]{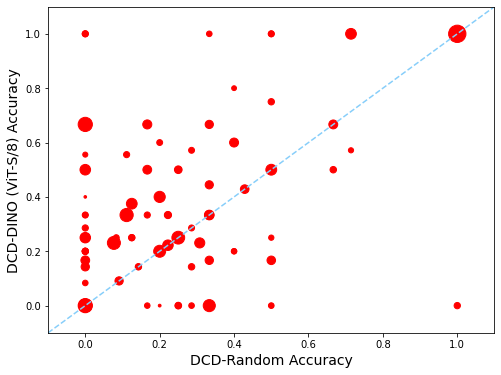}
        \vspace{0.05cm}
        \caption{accuracy for 160 classes for DCD-DINO (Y-axis) and -Random (X-axis) on heldout set}
        \label{fig:heldout_160}
        \end{subfigure} \qquad
        \begin{subfigure}{0.3\linewidth}
        \includegraphics[trim=0cm 0cm 0cm 0cm, clip, width=\linewidth]{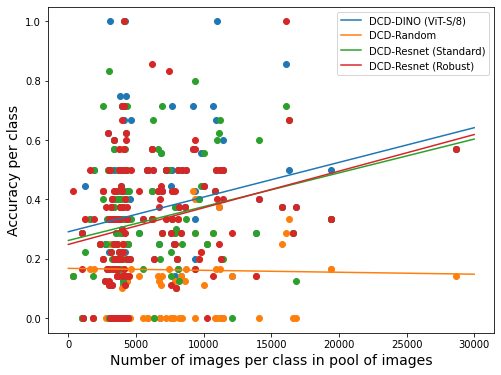}
        \vspace{-0.05cm}
        \caption{heldout set accuracy for each class $i \in \errorclasses$ (Y-axis) as $|\flickr(i)|$ increases (X-axis)}
        \label{fig:models_160}
        \end{subfigure}
        \vspace{-0.1cm}
        \caption{In (a) and (b), we plot DCD-DINO accuracy (Y-axis) and DCD-Random (X-axis) for all 160 classes. The size of each dot is proportional to the number of images in $\flickr$ for that class ($|\flickr(i)|$) and dashed line represents $Y=X$. In (c), we plot the accuracy per class on the heldout set as $|\flickr(i)|$ increases. We also show the corresponding linear regression lines.}
    \label{fig:bubble}
\end{figure*}



\subsection{Baseline models}\label{sec:baseline_models}
\noindent We compare against three baseline models:  

\begin{itemize}[leftmargin=*]
\item \textbf{original}: trained on the ImageNet training set (no additional training images are added)

\item  \textbf{DCD-Complete}: We add the complete dataset $\flickr$ of size $952,022$ (Section \ref{sec:remove_test_set}). Since this leads to a disproportionately large number of images from the $160$ classes ($\errorclasses$) in $\flickr$, we assign weights $(w(i))$ to these classes so that the total weight for each class is the same as in ImageNet: 
$$\forall\ i \in \errorclasses,\ \  w(i) \times |\imagenet_{train}(i) \cup \flickr(i)| = |\imagenet_{train}(i)|$$ 
\item  \textbf{DCD-Random}: For each class $i \in \errorclasses$, we randomly select $46 \times |\dist_{seed}(i)|$ images (without replacement) from $\flickr(i)$. Note that this set has the same size as our debug-train set. This ensures a fair comparison across models. From Table \ref{table:statistics_single_model}, the dataset has size $1031 \times 46 = 47426$.
\end{itemize}

\subsection{Table details}
\noindent In Table \ref{table:single_model}, we show results on different test sets namely, \\
\textbf{MFreq}: only the ImageNet-V2 MatchedFrequency set \\
\textbf{Complement}: all ImageNet-V2 images except Mfreq\\
\textbf{INet}: standard ImageNet test set. \\
Details for ``MFreq'' and ``Complement'' are in Section \ref{sec:seed_heldout_sets}. 

In ``Accuracy on different sets'', we show the model accuracy on various subsets of the test sets.  In ``160 classes'', we show the accuracy on images from classes in $\errorclasses$ and ``1000 classes'': from all 1000 classes in ImageNet. In ``incorrectly classified'', accuracy on images (from 160 class subset) that are incorrectly classified by the ``original'' model.

In the last four rows, ``Debug-Train method'' denotes the model $\Phi$ used for computing visual similarity distances. DCD-DINO (ViT-S/8 and ViT-S/16) denote the models trained using DINO ViT-S/8 and ViT-S/16 features. DCD-Resnet (Standard and Robust) denote the models trained using Standard and Robust Resnet-50 features.

\subsection{Discussion}
\noindent \textbf{adding the complete set}: While one may believe that noisy data is better than no data, we observe that naively adding noisy data ($\flickr$) has diminishing returns. In ``DCD-Complete'', 952,022 extra images are added while in ``DCD-Random'' only 47,426. Even though we add 20 $\times$ more images in ``Complete'', accuracy on seed and heldout are only marginally better. In fact, under other metrics, such as ``160 classes'' (INet), accuracy for ``Complete'' is $49.22\%$  significantly below ``original'' $62.89\%$ ($-13.67\%$) and ``Random'' $63.91\%$ ($-14.69\%$). This suggests that naively adding large amounts of noisy data can hurt model performance. 

\noindent \textbf{comparing models}: We want to compare the quality of image embeddings for visual similarity matching obtained using different models. We observe that both DINO models achieve significantly higher accuracy on ``Seed'' compared to the Resnet models: DINO ViT-S/8 achieves 36.28\% compared to 33.07\% for Robust Resnet-50. This provides evidence that DINO models are better suited for similarity matching. Similar trend is also observed for the Heldout set. However, between Standard and Robust Resnet-50, the results on ``Seed'' are comparable suggesting that adversarial robustness is not critical for similarity matching.

\noindent \textbf{comparing baselines and our method}: We observe that DINO ViT-S/8 achieves significantly improved results compared to both ``DCD-complete'' and ``DCD-random''. On ``Heldout'', we achieve 29.62\% compared to 23.09\% for the next best i.e. gain of 6.53\%. On the ``Complement'' set (160 classes), we achieve 63.85\%: gain of 5.04\% compared to 58.81\% for the next best model. On the 1000 classes sets, we achieve slightly improved results on all sets: 76.42\% on ``Complement'', compared to 76.12\% (+0.3\%). Similarly on INet, we achieve 76.54\% similar to 76.47\%. The model performance is maintained on all the test sets. 

\noindent \textbf{accuracy on different classes}: We now compare the seed and heldout accuracy for different classes $i \in \errorclasses$ as $|\flickr(i)|$ varies. In both Figures \ref{fig:seed_160} and \ref{fig:heldout_160}, we observe that DCD-DINO achieves better accuracy on most classes comparing to DCD-Random (as most points lie above the dashed y=x line). Also, DCD-DINO achieves better accuracy for classes with larger amount of data (larger red dots). In Figure \ref{fig:models_160}, we compare accuracy per class as $|\flickr(i)|$ increases (x-axis) for four different models: DCD-Resnet (Standard and Robust), DCD-DINO (ViT-S/8) and DCD-Random. We see that as $|\flickr(i)|$ increases, we achieve better accuracy (according to the linear regression lines) for all methods except DCD-Random. For DCD-Random, there is a slight accuracy reduction for large $|\flickr(i)|$. Our results provide evidence that by obtaining large amounts of weakly-labeled data and adding selected images to the training sets, we can achieve significantly improved results. 

\section{Related work}\label{sec:related_work}
\noindent\textbf{Dataset Design}: Several previous works \cite{he2016deep, krizhevsky2017imagenet, dosovitskiy2020image} use gigantic amounts of training data to achieve high performance. For example, models such as Basic \cite{pham2021combined} and CLIP \cite{radford2021learning} use web-scale datasets of sizes 6.6 billion and 400 million images respectively. The availability of large weakly labeled web data combined with self-supervision methods \cite{brown2020language, chen2020simple, devlin2018bert, he2020momentum, he2022masked} has made it easier to train such models. However, recent work \cite{nguyen2022quality, yang2022deep} suggests that adding targeted training images may be more effective. \\
\noindent\textbf{Debugging and Explainability}: Most of the existing works on explainability of deep networks focus on inspecting the decisions for a single image \cite{ZeilerF13, MahendranV15, DosovitskiyB15, YosinskiCNFL15, NguyenYC16, sanitychecks2018, zhou2018interpreting, chang2018explaining, olah2018the, Yeh2019OnT, carter2019activation, oshaughnessy2020generative, sturmfels2020visualizing, verma2020counterfactual}. These include saliency maps \cite{Simonyan2013DeepIC, Sundararajan2017AxiomaticAF, Smilkov2017SmoothGradRN, Singla2019UnderstandingIO}, activation maps \cite{ZhouKLOT15,SelvarajuDVCPB16, Bau30071, Ismail2019AttentionDL, Ismail2020BenchmarkingDL}, removing image patches \cite{jain2022missingness}, interpreting local decision boundaries \cite{ribeiro2016},  finding influential \cite{pangweikoh2021} or counterfactual inputs \cite{NguyenYC14, mahendran15, deviparikhcounterfactual19}. 
However, examining several single-image explanations for debugging can be time-consuming. Therefore, several works focus on identifying failure modes across a large set of images\cite{pandora1809, zhang2018manifold, chung2019slice, wu2019errudite, leclerc20213db, sparsewong21b, singlaCVPR2021, salientimagenet2021} or introducing datasets to stress test model performance on images with the main objects in uncommon or rare settings \cite{objectnet19, kattakinda22a, moayeri2022hard, hendrycks2021nae}. Another class of works focus on making edits to the model to modify its predictions on specific inputs \cite{santurkar2021editing, mitchell2021fast, decao2021editing}. 

\noindent \textbf{Visual Similarity metrics}: Several prior similarity metrics have been proposed in the literature \cite{1284395, 5705575, 1292216}. The state-of-the-art LPIPS metric \cite{zhang2018unreasonable} concatenates intermediate features of a deep network to compute the embeddings. 

\section{Acknowledgements}
This project was supported in part by NSF CAREER AWARD 1942230, HR001119S0026 (GARD), ONR YIP award N00014-22-1-2271, Army Grant No. W911NF2120076, the NSF award CCF2212458, Meta grant 23010098.

{\small
\bibliographystyle{ieee_fullname}
\bibliography{egbib}
}

\clearpage

\onecolumn

\noindent {\bf \LARGE Appendix}
\appendix

\section{Selecting 160 classes for model debugging}\label{sec:appendix_160_classes}
We first selected $30$ models with different architectures from the timm library. 
From these $30$ models, we selected $20$ models that achieved highest accuracy on the ImageNet-V2 set. The $20$ models were:
swin\_base\_patch4\_window7\_224, swin\_small\_patch4\_window7\_224,
convit\_base, deit\_base\_patch16\_224, convit\_small,
swin\_tiny\_patch4\_window7\_224, resnet50d, mixnet\_xl, seresnet50,
deit\_small\_patch16\_224, resnext50\_32x4d, efficientnet\_b4, resnet50,
efficientnet\_b3, wide\_resnet101\_2, efficientnet\_b0,
vit\_base\_patch16\_224, resnet34, mnasnet\_a1, vit\_small\_patch16\_224.

ImageNet-V2 consists of $3$ test sets namely: MatchedFrequency, TopImages and Threshold0.7. We used the MatchedFrequency version because models achieve the lowest accuracy on this set (making it suitable for debugging). Next, we selected classes with at least $3$ images on which all $20$ selected models were inaccurate on the MatchedFrequency set. 

This resulted in total $160$ classes. 

\section{Set of models used in the multiple-model setting}\label{sec:appendix_all_model}
We used the same models as in Section \ref{sec:appendix_160_classes} except that the Resnet-$50$ model included was not from the timm library but another Resnet-$50$ trained from scratch by us. We used the same Resnet-$50$ as the one used in the single-model setting as this makes the comparison between multiple-model and single-model settings easier. This again resulted in total $20$ models.

\section{Models used for visual similarity matching}\label{sec:appendix_similarity_matching}
We experiment with four pretrained models $\Phi$ for computing these distances: Standard Resnet-50, Robust Resnet-50, DINO ViT-S/16 and DINO ViT-S/8 \cite{dino}. Here, Standard Resnet-50 model is the original trained model we are trying to debug. We use Robust Resnet-50 because adversarially robust models have the unique property that if you try to optimize two visually different images to minimize the visual similarity distance between them (using the robust model as the feature extractor), the resulting images look visually very similar. The same is not true for standard models as shown in  \cite{Engstrom2019LearningPR}. This suggests that robust models may be better suited for visual similarity matching and we investigate this in the paper. We selected the DINO ViT-S/16 and ViT-S/8 models because the penultimate layer features for these models are known to be good kNN classifiers \cite{dino} achieving 74.5\% and 78.3\% accuracy on ImageNet respectively.

\section{Removing images ``visually similar'' to test-sets}\label{sec:appendix_remove_test_set}
Since the model performance on test set can be trivially improved by adding images from the test set to the training set, it is critically important to ensure that the new images added to the training set are ``sufficiently different'' from the test set images. To this end, we rely on the representation of a Robust Resnet-50 model $\Phi_{r}$ (for two images $\bx, \bz$, $\|\Phi_{r}(\bx) - \Phi_{r}(\bz)\|^{2}$ is the visual similarity metric). Using this metric, we argue that for each class, the newly added images should be at least as different from test set images as they are between the ImageNet train/test sets. 

Let $\imagenet_{train}$ and $\imagenet_{test}$ denote the ImageNet train and test sets. Thus, for each label $i \in \errorclasses$, we first compute a threshold visual similarity distance $\tau(i)$ using ImageNet as follows:
\begin{align*}
    &\tau(i) = \min \|\Phi_{r}(\bx) - \Phi_{r}(\bz)\|^{2}\\ 
    &\text{where } \bx \in \imagenet_{train}(i),\ \ \bz \in \imagenet_{test}(i)
\end{align*}
For each class $i$, $\tau(i)$ captures the minimum perceptual distance that should exist between the train and test images for class $i$. Let $\union$ denote the union of all test sets 
that we want to evaluate our model on. This includes the seed set, heldout set and every test set on which we want to maintain model performance. We want to select the images $\bx \in \flickr(i)$ that have visual distance $> \tau(i)$,\ from all images $\bz \in \union(i)$. To this end, we construct a new dataset $\flickr$ as follows: 
\begin{align*}
& \flickr(i) = \{\bx \in \bar{\flickr}(i) : \min_{\bz\ \in\  \union(i)} \|\Phi_{r}(\bx) - \Phi_{r}(\bz)\|^{2} > \tau(i) \} 
\end{align*}

\begin{figure}[h!]
\centering
\begin{subfigure}{0.15\textwidth}
  \includegraphics[trim=0cm 0cm 0cm 0.9cm, clip, width=\linewidth]{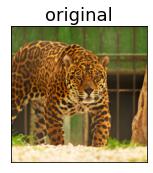}
 \caption{original image}
\end{subfigure}\ 
\begin{subfigure}{0.15\textwidth}
  \includegraphics[trim=0cm 0cm 0cm 0.9cm, clip, width=\linewidth]{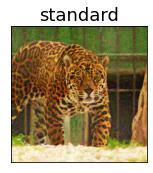}
\caption{standard model}
\end{subfigure}\ 
\begin{subfigure}{0.15\textwidth}
  \includegraphics[trim=0cm 0cm 0cm 0.9cm, clip, width=\linewidth]{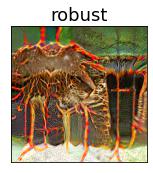}
\caption{robust model}
\end{subfigure}\ \ 
\caption{(a) initial image (denoted by $\bx$). We solve $\bz = \argmax_{\bz: \|\bz - \bx\| \leq \rho} \|\Phi(\bz) - \Phi(\bx)\|^{2}$ using $\Phi$ as Standard Resnet-50 in (b) and Robust Resnet-50 in (c). In (b), $\bx$ and $\bz$ look identical to a human eye. In (c), $\bx$ and $\bz$ are visually very different.}
\label{fig:make_diff}
\end{figure}

We use the Robust Resnet-50 model for removing similar images because for a standard model, using gradient based adversarial attacks, it is possible to construct pairs of images that look exactly the same to the human eye, yet map to different representations. However, for a robust model, such adversarial attacks lead to changes that are visible to the human eye and the resulting pairs of images are visually different. Thus, even if $\|\Phi(\bx) - \Phi(\bz)\|^{2}$ is very large for a standard model, $\bx$ and $\bz$ may still look identical to a human. This is further illustrated in Figure \ref{fig:make_diff}. Thus suggests that using a robust model leads to a more reliable metric for removing images that are similar to some reference image. 


\section{Algorithm for constructing debug-validation and debug-train sets}\label{sec:appendix_sec_algo_non_disjoint}

To avoid overlaps when selecting ``visually similar'' images per image, we use the below procedure: 
\begin{itemize}
\item For each $\bx \in \dist_{seed}(i)$, we initialize the set: $\near[\bx] = \{\}$. 
\item For all pairs $\bx \in \dist_{seed}(i),\ \bz \in \flickr(i)$, we compute the distance $\|\Phi(\bx) - \Phi(\bz)\|^{2}$. This gives the matrix $\mathrm{D}$.
\item Let $\bx^{*},\ \bz^{*}$ be the pair with the minimum distance in $\mathrm{D}$. We add $\bz^{*}$ to $\near[\bx^{*}]$ and set $\mathrm{D}[:, \bz^{*}] = \infty$ (this prevents $\bz^{*}$ from being selected again, thus no overlaps).
\item If the number of elements in $\near[\bx^{*}]$ equals to $k$, we set $\mathrm{D}[\bx^{*}, :] = \infty$ (this prevents $\bx^{*}$ from being selected again if $\near[\bx^{*}]$ contains $k$ elements).
\item We repeat this procedure until $\forall\ \bx$, $\near[\bx]$ is of size $k$. 
\end{itemize}
This is equivalent to Algorithm \ref{alg:disjoint_sets} with $\query = \dist_{seed}(i), \web = \flickr(i)$. 
The complete de-val set is given by: 
\begin{align}
\val(i) = \cup_{\bx \in \dist_{seed}(i)}\ \near[\bx] \label{eq:val_set}
\end{align}
Similarly, the debug-train procedure is constructed using the inputs  $\query = \dist_{seed}(i), \web = (\flickr - \val)(i)$ to Algorithm \ref{alg:disjoint_sets}.

\begin{algorithm}[H]
\SetAlgoLined
\SetKwInput{KwInput}{Input}                
\SetKwInput{KwOutput}{Output}              
\SetKwInput{KwReturn}{Return}              
\DontPrintSemicolon
\KwInput{seed set: $\query$,\ web data: $\web$,\ count: $k$}
\KwOutput{$\near$}
\For{$\bx \in \query$}
{
    $\near[\bx] \gets \{\}$\\
    \For{$\bz \in \web$}
    {
        $\mathrm{D}[\bx,\ \bz] \gets \|\Phi(\bx) - \Phi(\bz)\|^{2}$\\
    }
}
\While{$\exists\ \bx:\ \left|\near[\bx]\right| < k$}
{
    $\bx,\ \bz = \arg\min \mathrm{D}$ \\ 
    $\near[\bx] \gets \near[\bx] \cup \bz$\\
    $\mathrm{D}[:,\ \bz] \gets \infty $\\
    \If {$\left|\near[\bx]\right| == k$} 
    {
        $\mathrm{D}[\bx,\ :] \gets \infty $
    }
}
\caption{Finding disjoint sets of similar images}
\label{alg:disjoint_sets}
\end{algorithm}

\section{Results in the multiple-model setting}\label{sec:appendix_multiple_model}

The seed and heldout sets contain 563 and 264 images respectively.



\begin{table*}[h!]
    \centering
    \renewcommand{\arraystretch}{1.3} 
    \centering
    \begin{tabular}{l | ll | lll |  lll }
    \toprule
    \multirow{3}{1.5cm}{\textbf{Debug-Train method}} & \multicolumn{8}{|c}{\textbf{Accuracy on different sets}} \\ 
    \cmidrule{2-9}
       & \multicolumn{2}{|c|}{\textbf{incorrectly classified}} 
      & \multicolumn{3}{c|}{\textbf{subset of 160 classes}} & \multicolumn{3}{c}{\textbf{all 1000 classes}} \\ 
     & Seed & Heldout & MFreq & Compl. & INet &  MFreq & Compl. & INet \\
\midrule
original & 0\% & 0\% & 35.56\% & 56.89\% & 62.89\% &  63.70\% & 76.12\% & 76.47\%  \\
DCD-Complete & 8.88\% & 7.95\% & 37.56\% & 53.84\% & 49.22\% & 61.70\% & 72.93\% & 75.07\%  \\
DCD-Random & 2.66\% & 3.03\% & 40.31\% & 59.35\% & 63.95\% & 63.76\% & 75.96\% & 76.60\%  \\
\midrule
DCD-DINO (ViT-S/8) & 19.89\% & \textbf{8.33\%} & 47.31\% & 60.25\% & 63.74\% & 65.32\% & 75.81\% & 76.61\% \\
DCD-DINO (ViT-S/16)  & \textbf{21.67\%} & 4.92\% & \textbf{47.38\%} & 60.43\% & 63.94\% & 64.93\% & 76.03\% & 76.49\% \\
DCD-Resnet (Standard) & 17.58\% & 7.57\% & 46.56\% & \textbf{61.27\%} & \textbf{64.34\%} & 64.69\% & \textbf{76.23\%} & \textbf{76.63\%} \\
DCD-Resnet (Robust)  & 15.98\% & 6.06\% & 45.31\% & 60.43\% & 64.00\% & 64.95\% & 75.80\% & 76.58\% \\
\bottomrule
\end{tabular}
\caption{Results for \emph{multiple-model} seed and heldout sets. ``INet'' denotes the ImageNet test set, ``MFreq'' denotes the ImageNet-V2\cite{pmlr-v97-recht19a} MatchedFrequency set, ``Compl.'' (Complement) denotes the set of all ImageNet-V2 images excluding ``MFreq'' images. }
\label{table:multiple_model}
\end{table*}

\subsection{Table details}
In Table \ref{table:multiple_model}, the columns ``Seed'' and ``Heldout'' show the accuracy on images in ``Mfreq'' and ``Complement'' sets respectively from the 160 classes ($\errorclasses$) that were incorrectly classified by the all 20 models (\ref{sec:appendix_all_model}). Note that these are equivalent to the seed and heldout sets discussed in Section \ref{sec:seed_heldout_sets}.

In the last four rows, ``Debug-Train method'' denotes the model $\Phi$ used for computing visual similarity distances. DCD-DINO (ViT-S/8) and (ViT-S/16) denote the models trained using DINO ViT-S/8 and ViT-S/16 models. DCD-Resnet (Standard) and (Robust) denote the models trained using Standard and Robust Resnet-50 models. 

\subsection{Discussion}

For the ``Debug-Train method'': DCD-DINO (ViT-S/8), in the column ``incorrectly classified'' achieves the highest accuracy on Heldout: $8.33\%$. Notably, on heldout, the accuracy is only slightly higher than ``DCD-Complete'' (0.38\%). However, ``DCD-Complete'' shows 13.67\% drop in performance on ImageNet (160 class subset). Although the accuracy of DCD-Random is similar to original model on ``INet (160)'', it performs considerable worse compared to DCD-DINO (ViT-S/8) on Heldout: we achieve 8.33\% significantly better the ``DCD-Random'' method (3.03\% i.e. gain of 5.3\%). 

While one may believe that the images misclassified by 20 ImageNet trained models would have multiple objects or may be mislabeled. We find that this is not the case for several images from both the seed and heldout-debug sets. We show images correctly classified by our model and misclassified by 20 ImageNet trained models in Appendix \ref{sec:seed_set_multiple_model_examples} (for the seed-debug set) and Appendix \ref{sec:heldout_set_multiple_model_examples} (for the heldout-debug set).



\clearpage

\clearpage

\onecolumn
\section{Examples of images from the seed-debug set}\label{sec:seed_set_multiple_model_examples}
\begin{figure*}[h!]
\centering
\begin{subfigure}{0.19\textwidth}
  \includegraphics[trim=0cm 0cm 0cm 0cm, clip, width=\linewidth]{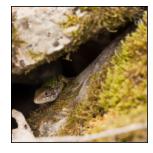}
  \caption*{green lizard}
\end{subfigure}\ \ 
\begin{subfigure}{0.19\textwidth}
  \includegraphics[trim=0cm 0cm 0cm 0cm, clip, width=\linewidth]{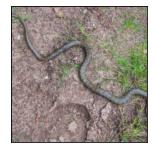}
  \caption*{garter snake}
\end{subfigure}\ \ 
\begin{subfigure}{0.19\textwidth}
  \includegraphics[trim=0cm 0cm 0cm 0cm, clip, width=\linewidth]{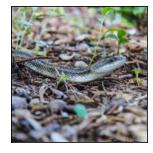}
  \caption*{water snake}
\end{subfigure}\ \ 
\begin{subfigure}{0.19\textwidth}
  \includegraphics[trim=0cm 0cm 0cm 0cm, clip, width=\linewidth]{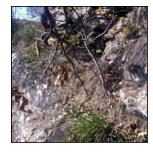}
  \caption*{diamondback}
\end{subfigure}\ \ 
\begin{subfigure}{0.19\textwidth}
  \includegraphics[trim=0cm 0cm 0cm 0cm, clip, width=\linewidth]{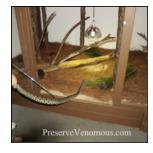}
  \caption*{diamondback}
\end{subfigure} \\
\begin{subfigure}{0.19\textwidth}
  \includegraphics[trim=0cm 0cm 0cm 0cm, clip, width=\linewidth]{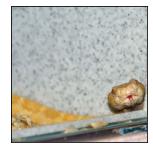}
  \caption*{diamondback}
\end{subfigure}\ \ 
\begin{subfigure}{0.19\textwidth}
  \includegraphics[trim=0cm 0cm 0cm 0cm, clip, width=\linewidth]{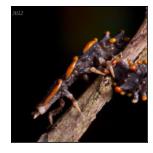}
  \caption*{trilobite}
\end{subfigure}\ \ 
\begin{subfigure}{0.19\textwidth}
  \includegraphics[trim=0cm 0cm 0cm 0cm, clip, width=\linewidth]{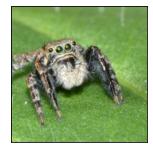}
  \caption*{garden spider}
\end{subfigure}\ \ 
\begin{subfigure}{0.19\textwidth}
  \includegraphics[trim=0cm 0cm 0cm 0cm, clip, width=\linewidth]{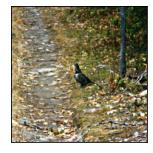}
  \caption*{black grouse}
\end{subfigure}\ \ 
\begin{subfigure}{0.19\textwidth}
  \includegraphics[trim=0cm 0cm 0cm 0cm, clip, width=\linewidth]{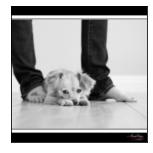}
  \caption*{papillon}
\end{subfigure} \\
\begin{subfigure}{0.19\textwidth}
  \includegraphics[trim=0cm 0cm 0cm 0cm, clip, width=\linewidth]{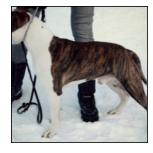}
  \caption*{american staffordshire terrier}
\end{subfigure}\ \ 
\begin{subfigure}{0.19\textwidth}
  \includegraphics[trim=0cm 0cm 0cm 0cm, clip, width=\linewidth]{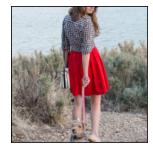}
  \caption*{wire-haired fox terrier}
\end{subfigure}\ \ 
\begin{subfigure}{0.19\textwidth}
  \includegraphics[trim=0cm 0cm 0cm 0cm, clip, width=\linewidth]{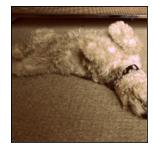}
  \caption*{wire-haired fox terrier}
\end{subfigure}\ \ 
\begin{subfigure}{0.19\textwidth}
  \includegraphics[trim=0cm 0cm 0cm 0cm, clip, width=\linewidth]{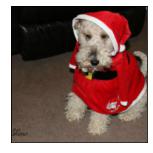}
  \caption*{wire-haired fox terrier}
\end{subfigure}\ \ 
\begin{subfigure}{0.19\textwidth}
  \includegraphics[trim=0cm 0cm 0cm 0cm, clip, width=\linewidth]{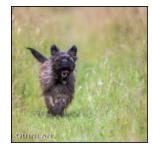}
  \caption*{lakeland terrier}
\end{subfigure} \\
\begin{subfigure}{0.19\textwidth}
  \includegraphics[trim=0cm 0cm 0cm 0cm, clip, width=\linewidth]{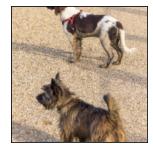}
  \caption*{lakeland terrier}
\end{subfigure}\ \ 
\begin{subfigure}{0.19\textwidth}
  \includegraphics[trim=0cm 0cm 0cm 0cm, clip, width=\linewidth]{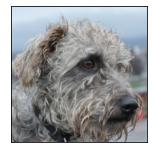}
  \caption*{lakeland terrier}
\end{subfigure}\ \ 
\begin{subfigure}{0.19\textwidth}
  \includegraphics[trim=0cm 0cm 0cm 0cm, clip, width=\linewidth]{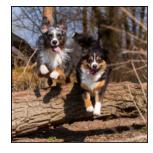}
  \caption*{australian terrier}
\end{subfigure}\ \ 
\begin{subfigure}{0.19\textwidth}
  \includegraphics[trim=0cm 0cm 0cm 0cm, clip, width=\linewidth]{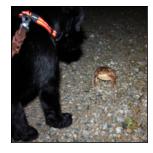}
  \caption*{standard schnauzer}
\end{subfigure}\ \ 
\begin{subfigure}{0.19\textwidth}
  \includegraphics[trim=0cm 0cm 0cm 0cm, clip, width=\linewidth]{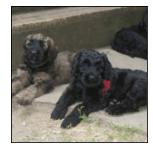}
  \caption*{briard}
\end{subfigure} \\
\begin{subfigure}{0.19\textwidth}
  \includegraphics[trim=0cm 0cm 0cm 0cm, clip, width=\linewidth]{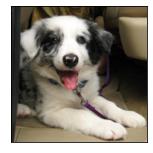}
  \caption*{collie}
\end{subfigure}\ \ 
\begin{subfigure}{0.19\textwidth}
  \includegraphics[trim=0cm 0cm 0cm 0cm, clip, width=\linewidth]{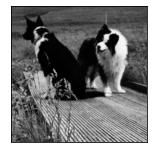}
  \caption*{collie}
\end{subfigure}\ \ 
\begin{subfigure}{0.19\textwidth}
  \includegraphics[trim=0cm 0cm 0cm 0cm, clip, width=\linewidth]{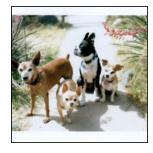}
  \caption*{miniature pinscher}
\end{subfigure}\ \ 
\begin{subfigure}{0.19\textwidth}
  \includegraphics[trim=0cm 0cm 0cm 0cm, clip, width=\linewidth]{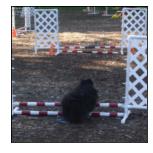}
  \caption*{pomeranian}
\end{subfigure}\ \ 
\begin{subfigure}{0.19\textwidth}
  \includegraphics[trim=0cm 0cm 0cm 0cm, clip, width=\linewidth]{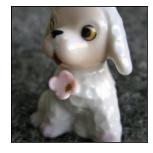}
  \caption*{toy poodle}
\end{subfigure} 
\caption{Examples of images (from the seed-debug set) on which $20$ highly accurate ImageNet trained models predict incorrectly (ground truth label below each image). However, models trained using our framework make correct predictions on all of them.}
\label{fig:appendix_images_0_25}
\end{figure*}

\begin{figure*}[h!]
\centering
\begin{subfigure}{0.19\textwidth}
  \includegraphics[trim=0cm 0cm 0cm 0cm, clip, width=\linewidth]{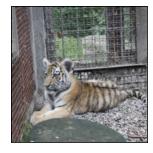}
  \caption*{tiger cat}
\end{subfigure}\ \ 
\begin{subfigure}{0.19\textwidth}
  \includegraphics[trim=0cm 0cm 0cm 0cm, clip, width=\linewidth]{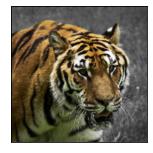}
  \caption*{tiger cat}
\end{subfigure}\ \ 
\begin{subfigure}{0.19\textwidth}
  \includegraphics[trim=0cm 0cm 0cm 0cm, clip, width=\linewidth]{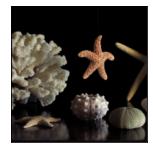}
  \caption*{sea urchin}
\end{subfigure}\ \ 
\begin{subfigure}{0.19\textwidth}
  \includegraphics[trim=0cm 0cm 0cm 0cm, clip, width=\linewidth]{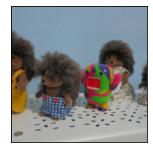}
  \caption*{porcupine}
\end{subfigure}\ \ 
\begin{subfigure}{0.19\textwidth}
  \includegraphics[trim=0cm 0cm 0cm 0cm, clip, width=\linewidth]{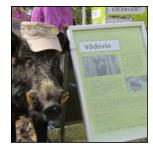}
  \caption*{hog}
\end{subfigure} \\
\begin{subfigure}{0.19\textwidth}
  \includegraphics[trim=0cm 0cm 0cm 0cm, clip, width=\linewidth]{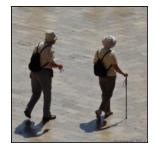}
  \caption*{backpack}
\end{subfigure}\ \ 
\begin{subfigure}{0.19\textwidth}
  \includegraphics[trim=0cm 0cm 0cm 0cm, clip, width=\linewidth]{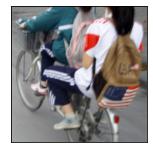}
  \caption*{backpack}
\end{subfigure}\ \ 
\begin{subfigure}{0.19\textwidth}
  \includegraphics[trim=0cm 0cm 0cm 0cm, clip, width=\linewidth]{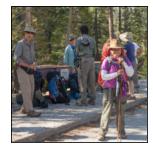}
  \caption*{backpack}
\end{subfigure}\ \ 
\begin{subfigure}{0.19\textwidth}
  \includegraphics[trim=0cm 0cm 0cm 0cm, clip, width=\linewidth]{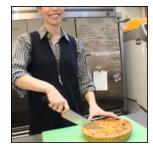}
  \caption*{bakery}
\end{subfigure}\ \ 
\begin{subfigure}{0.19\textwidth}
  \includegraphics[trim=0cm 0cm 0cm 0cm, clip, width=\linewidth]{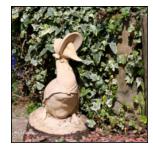}
  \caption*{bonnet}
\end{subfigure} \\
\begin{subfigure}{0.19\textwidth}
  \includegraphics[trim=0cm 0cm 0cm 0cm, clip, width=\linewidth]{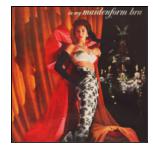}
  \caption*{brassiere}
\end{subfigure}\ \ 
\begin{subfigure}{0.19\textwidth}
  \includegraphics[trim=0cm 0cm 0cm 0cm, clip, width=\linewidth]{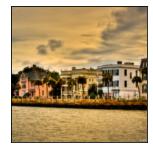}
  \caption*{breakwater}
\end{subfigure}\ \ 
\begin{subfigure}{0.19\textwidth}
  \includegraphics[trim=0cm 0cm 0cm 0cm, clip, width=\linewidth]{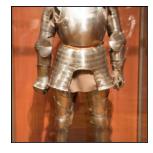}
  \caption*{breastplate}
\end{subfigure}\ \ 
\begin{subfigure}{0.19\textwidth}
  \includegraphics[trim=0cm 0cm 0cm 0cm, clip, width=\linewidth]{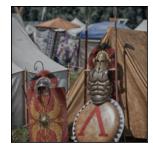}
  \caption*{breastplate}
\end{subfigure}\ \ 
\begin{subfigure}{0.19\textwidth}
  \includegraphics[trim=0cm 0cm 0cm 0cm, clip, width=\linewidth]{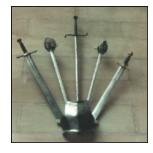}
  \caption*{breastplate}
\end{subfigure} \\
\begin{subfigure}{0.19\textwidth}
  \includegraphics[trim=0cm 0cm 0cm 0cm, clip, width=\linewidth]{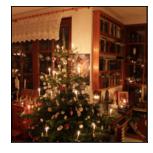}
  \caption*{candle}
\end{subfigure}\ \ 
\begin{subfigure}{0.19\textwidth}
  \includegraphics[trim=0cm 0cm 0cm 0cm, clip, width=\linewidth]{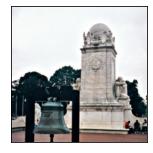}
  \caption*{chime}
\end{subfigure}\ \ 
\begin{subfigure}{0.19\textwidth}
  \includegraphics[trim=0cm 0cm 0cm 0cm, clip, width=\linewidth]{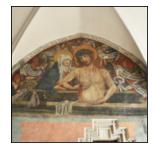}
  \caption*{church}
\end{subfigure}\ \ 
\begin{subfigure}{0.19\textwidth}
  \includegraphics[trim=0cm 0cm 0cm 0cm, clip, width=\linewidth]{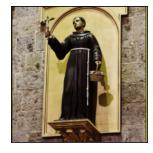}
  \caption*{church}
\end{subfigure}\ \ 
\begin{subfigure}{0.19\textwidth}
  \includegraphics[trim=0cm 0cm 0cm 0cm, clip, width=\linewidth]{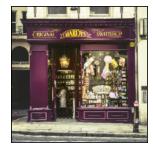}
  \caption*{confectionery}
\end{subfigure} \\
\begin{subfigure}{0.19\textwidth}
  \includegraphics[trim=0cm 0cm 0cm 0cm, clip, width=\linewidth]{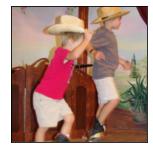}
  \caption*{cowboy boot}
\end{subfigure}\ \ 
\begin{subfigure}{0.19\textwidth}
  \includegraphics[trim=0cm 0cm 0cm 0cm, clip, width=\linewidth]{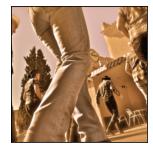}
  \caption*{cowboy boot}
\end{subfigure}\ \ 
\begin{subfigure}{0.19\textwidth}
  \includegraphics[trim=0cm 0cm 0cm 0cm, clip, width=\linewidth]{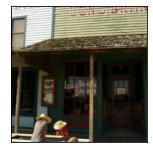}
  \caption*{cowboy boot}
\end{subfigure}\ \ 
\begin{subfigure}{0.19\textwidth}
  \includegraphics[trim=0cm 0cm 0cm 0cm, clip, width=\linewidth]{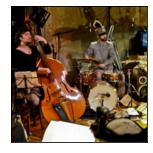}
  \caption*{drum}
\end{subfigure}\ \ 
\begin{subfigure}{0.19\textwidth}
  \includegraphics[trim=0cm 0cm 0cm 0cm, clip, width=\linewidth]{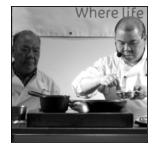}
  \caption*{frying pan}
\end{subfigure} 
\caption{Examples of images (from the seed-debug set) on which $20$ highly accurate ImageNet trained models predict incorrectly (ground truth label below each image). However, models trained using our framework make correct predictions on all of them.}
\label{fig:appendix_images_25_50}
\end{figure*}

\begin{figure*}[h!]
\centering
\begin{subfigure}{0.19\textwidth}
  \includegraphics[trim=0cm 0cm 0cm 0cm, clip, width=\linewidth]{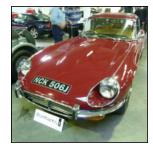}
  \caption*{grille}
\end{subfigure}\ \ 
\begin{subfigure}{0.19\textwidth}
  \includegraphics[trim=0cm 0cm 0cm 0cm, clip, width=\linewidth]{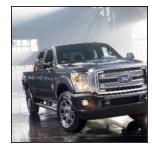}
  \caption*{grille}
\end{subfigure}\ \ 
\begin{subfigure}{0.19\textwidth}
  \includegraphics[trim=0cm 0cm 0cm 0cm, clip, width=\linewidth]{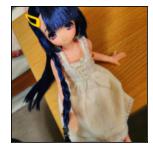}
  \caption*{hair slide}
\end{subfigure}\ \ 
\begin{subfigure}{0.19\textwidth}
  \includegraphics[trim=0cm 0cm 0cm 0cm, clip, width=\linewidth]{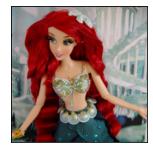}
  \caption*{hair slide}
\end{subfigure}\ \ 
\begin{subfigure}{0.19\textwidth}
  \includegraphics[trim=0cm 0cm 0cm 0cm, clip, width=\linewidth]{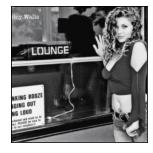}
  \caption*{jean}
\end{subfigure} \\
\begin{subfigure}{0.19\textwidth}
  \includegraphics[trim=0cm 0cm 0cm 0cm, clip, width=\linewidth]{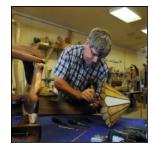}
  \caption*{lampshade}
\end{subfigure}\ \ 
\begin{subfigure}{0.19\textwidth}
  \includegraphics[trim=0cm 0cm 0cm 0cm, clip, width=\linewidth]{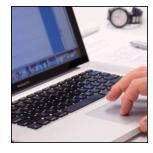}
  \caption*{laptop}
\end{subfigure}\ \ 
\begin{subfigure}{0.19\textwidth}
  \includegraphics[trim=0cm 0cm 0cm 0cm, clip, width=\linewidth]{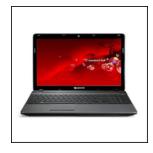}
  \caption*{laptop}
\end{subfigure}\ \ 
\begin{subfigure}{0.19\textwidth}
  \includegraphics[trim=0cm 0cm 0cm 0cm, clip, width=\linewidth]{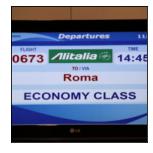}
  \caption*{monitor}
\end{subfigure}\ \ 
\begin{subfigure}{0.19\textwidth}
  \includegraphics[trim=0cm 0cm 0cm 0cm, clip, width=\linewidth]{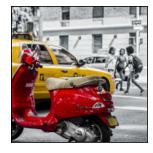}
  \caption*{moped}
\end{subfigure} \\
\begin{subfigure}{0.19\textwidth}
  \includegraphics[trim=0cm 0cm 0cm 0cm, clip, width=\linewidth]{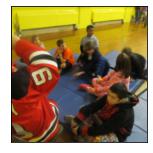}
  \caption*{pajama}
\end{subfigure}\ \ 
\begin{subfigure}{0.19\textwidth}
  \includegraphics[trim=0cm 0cm 0cm 0cm, clip, width=\linewidth]{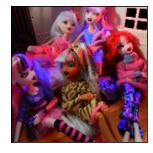}
  \caption*{pajama}
\end{subfigure}\ \ 
\begin{subfigure}{0.19\textwidth}
  \includegraphics[trim=0cm 0cm 0cm 0cm, clip, width=\linewidth]{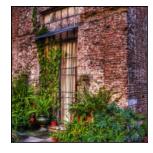}
  \caption*{patio}
\end{subfigure}\ \ 
\begin{subfigure}{0.19\textwidth}
  \includegraphics[trim=0cm 0cm 0cm 0cm, clip, width=\linewidth]{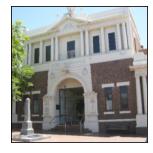}
  \caption*{pedestal}
\end{subfigure}\ \ 
\begin{subfigure}{0.19\textwidth}
  \includegraphics[trim=0cm 0cm 0cm 0cm, clip, width=\linewidth]{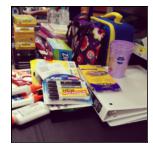}
  \caption*{pencil box}
\end{subfigure} \\
\begin{subfigure}{0.19\textwidth}
  \includegraphics[trim=0cm 0cm 0cm 0cm, clip, width=\linewidth]{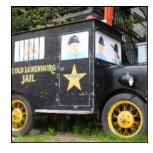}
  \caption*{police van}
\end{subfigure}\ \ 
\begin{subfigure}{0.19\textwidth}
  \includegraphics[trim=0cm 0cm 0cm 0cm, clip, width=\linewidth]{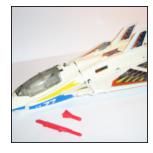}
  \caption*{projectile}
\end{subfigure}\ \ 
\begin{subfigure}{0.19\textwidth}
  \includegraphics[trim=0cm 0cm 0cm 0cm, clip, width=\linewidth]{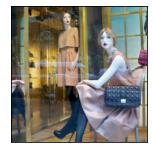}
  \caption*{purse}
\end{subfigure}\ \ 
\begin{subfigure}{0.19\textwidth}
  \includegraphics[trim=0cm 0cm 0cm 0cm, clip, width=\linewidth]{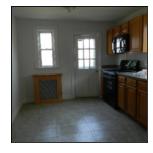}
  \caption*{radiator}
\end{subfigure}\ \ 
\begin{subfigure}{0.19\textwidth}
  \includegraphics[trim=0cm 0cm 0cm 0cm, clip, width=\linewidth]{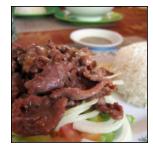}
  \caption*{restaurant}
\end{subfigure} \\
\begin{subfigure}{0.19\textwidth}
  \includegraphics[trim=0cm 0cm 0cm 0cm, clip, width=\linewidth]{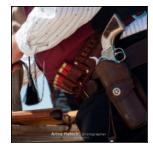}
  \caption*{revolver}
\end{subfigure}\ \ 
\begin{subfigure}{0.19\textwidth}
  \includegraphics[trim=0cm 0cm 0cm 0cm, clip, width=\linewidth]{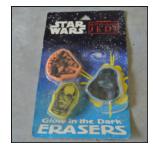}
  \caption*{rubber eraser}
\end{subfigure}\ \ 
\begin{subfigure}{0.19\textwidth}
  \includegraphics[trim=0cm 0cm 0cm 0cm, clip, width=\linewidth]{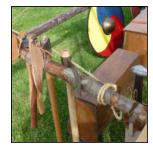}
  \caption*{scabbard}
\end{subfigure}\ \ 
\begin{subfigure}{0.19\textwidth}
  \includegraphics[trim=0cm 0cm 0cm 0cm, clip, width=\linewidth]{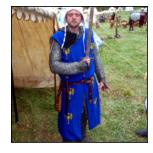}
  \caption*{scabbard}
\end{subfigure}\ \ 
\begin{subfigure}{0.19\textwidth}
  \includegraphics[trim=0cm 0cm 0cm 0cm, clip, width=\linewidth]{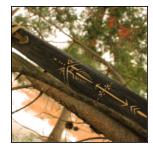}
  \caption*{scabbard}
\end{subfigure} 
\caption{Examples of images (from the seed-debug set) on which $20$ highly accurate ImageNet trained models predict incorrectly (ground truth label below each image). However, models trained using our framework make correct predictions on all of them.}
\label{fig:appendix_images_50_75}
\end{figure*}

\begin{figure*}[h!]
\centering
\begin{subfigure}{0.19\textwidth}
  \includegraphics[trim=0cm 0cm 0cm 0cm, clip, width=\linewidth]{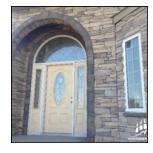}
  \caption*{stone wall}
\end{subfigure}\ \ 
\begin{subfigure}{0.19\textwidth}
  \includegraphics[trim=0cm 0cm 0cm 0cm, clip, width=\linewidth]{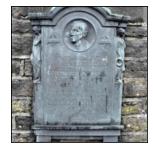}
  \caption*{stone wall}
\end{subfigure}\ \ 
\begin{subfigure}{0.19\textwidth}
  \includegraphics[trim=0cm 0cm 0cm 0cm, clip, width=\linewidth]{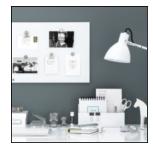}
  \caption*{table lamp}
\end{subfigure}\ \ 
\begin{subfigure}{0.19\textwidth}
  \includegraphics[trim=0cm 0cm 0cm 0cm, clip, width=\linewidth]{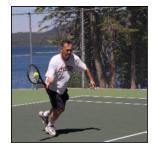}
  \caption*{tennis ball}
\end{subfigure}\ \ 
\begin{subfigure}{0.19\textwidth}
  \includegraphics[trim=0cm 0cm 0cm 0cm, clip, width=\linewidth]{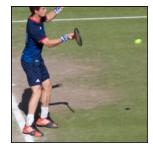}
  \caption*{tennis ball}
\end{subfigure} \\
\begin{subfigure}{0.19\textwidth}
  \includegraphics[trim=0cm 0cm 0cm 0cm, clip, width=\linewidth]{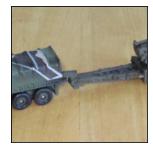}
  \caption*{tow truck}
\end{subfigure}\ \ 
\begin{subfigure}{0.19\textwidth}
  \includegraphics[trim=0cm 0cm 0cm 0cm, clip, width=\linewidth]{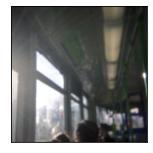}
  \caption*{window shade}
\end{subfigure}\ \ 
\begin{subfigure}{0.19\textwidth}
  \includegraphics[trim=0cm 0cm 0cm 0cm, clip, width=\linewidth]{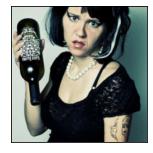}
  \caption*{wine bottle}
\end{subfigure}\ \ 
\begin{subfigure}{0.19\textwidth}
  \includegraphics[trim=0cm 0cm 0cm 0cm, clip, width=\linewidth]{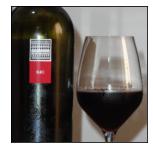}
  \caption*{wine bottle}
\end{subfigure}\ \ 
\begin{subfigure}{0.19\textwidth}
  \includegraphics[trim=0cm 0cm 0cm 0cm, clip, width=\linewidth]{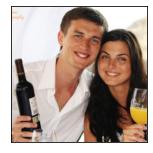}
  \caption*{wine bottle}
\end{subfigure} \\
\begin{subfigure}{0.19\textwidth}
  \includegraphics[trim=0cm 0cm 0cm 0cm, clip, width=\linewidth]{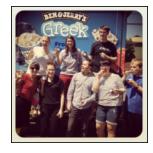}
  \caption*{ice cream}
\end{subfigure}\ \ 
\begin{subfigure}{0.19\textwidth}
  \includegraphics[trim=0cm 0cm 0cm 0cm, clip, width=\linewidth]{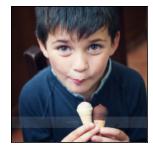}
  \caption*{ice cream}
\end{subfigure}\ \ 
\begin{subfigure}{0.19\textwidth}
  \includegraphics[trim=0cm 0cm 0cm 0cm, clip, width=\linewidth]{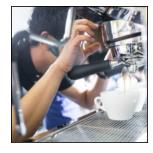}
  \caption*{espresso}
\end{subfigure}\ \ 
\begin{subfigure}{0.19\textwidth}
  \includegraphics[trim=0cm 0cm 0cm 0cm, clip, width=\linewidth]{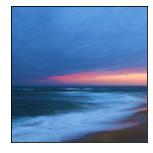}
  \caption*{promontory}
\end{subfigure}\ \ 
\begin{subfigure}{0.19\textwidth}
  \includegraphics[trim=0cm 0cm 0cm 0cm, clip, width=\linewidth]{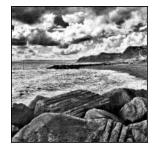}
  \caption*{seashore}
\end{subfigure} 
\caption{Examples of images (from the seed-debug set) on which $20$ highly accurate ImageNet trained models predict incorrectly (ground truth label below each image). However, models trained using our framework make correct predictions on all of them.}
\label{fig:appendix_images_75_90}
\end{figure*}

Figures \ref{fig:appendix_images_0_25}, \ref{fig:appendix_images_25_50}, \ref{fig:appendix_images_50_75} and \ref{fig:appendix_images_75_90} show several images from the seed-debug set on which we obtain correct predictions.

\clearpage

\clearpage

\clearpage

\clearpage
\section{Examples of images from the heldout-debug set}\label{sec:heldout_set_multiple_model_examples}
\begin{figure*}[h!]
\centering
\begin{subfigure}{0.19\textwidth}
  \includegraphics[trim=0cm 0cm 0cm 0cm, clip, width=\linewidth]{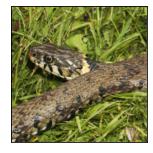}
  \caption*{garter snake}
\end{subfigure}\ \ 
\begin{subfigure}{0.19\textwidth}
  \includegraphics[trim=0cm 0cm 0cm 0cm, clip, width=\linewidth]{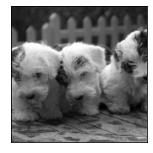}
  \caption*{dandie dinmont}
\end{subfigure}\ \ 
\begin{subfigure}{0.19\textwidth}
  \includegraphics[trim=0cm 0cm 0cm 0cm, clip, width=\linewidth]{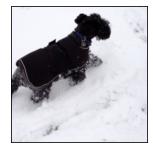}
  \caption*{standard schnauzer}
\end{subfigure}\ \ 
\begin{subfigure}{0.19\textwidth}
  \includegraphics[trim=0cm 0cm 0cm 0cm, clip, width=\linewidth]{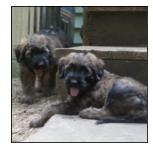}
  \caption*{briard}
\end{subfigure}\ \ 
\begin{subfigure}{0.19\textwidth}
  \includegraphics[trim=0cm 0cm 0cm 0cm, clip, width=\linewidth]{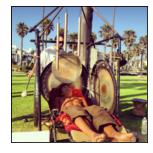}
  \caption*{chime}
\end{subfigure} \\
\begin{subfigure}{0.19\textwidth}
  \includegraphics[trim=0cm 0cm 0cm 0cm, clip, width=\linewidth]{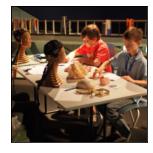}
  \caption*{dining table}
\end{subfigure}\ \ 
\begin{subfigure}{0.19\textwidth}
  \includegraphics[trim=0cm 0cm 0cm 0cm, clip, width=\linewidth]{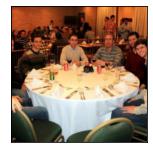}
  \caption*{dining table}
\end{subfigure}\ \ 
\begin{subfigure}{0.19\textwidth}
  \includegraphics[trim=0cm 0cm 0cm 0cm, clip, width=\linewidth]{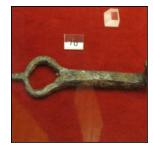}
  \caption*{hook}
\end{subfigure}\ \ 
\begin{subfigure}{0.19\textwidth}
  \includegraphics[trim=0cm 0cm 0cm 0cm, clip, width=\linewidth]{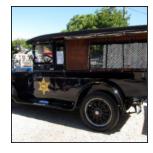}
  \caption*{police van}
\end{subfigure}\ \ 
\begin{subfigure}{0.19\textwidth}
  \includegraphics[trim=0cm 0cm 0cm 0cm, clip, width=\linewidth]{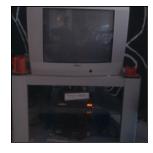}
  \caption*{screen}
\end{subfigure} \\
\begin{subfigure}{0.19\textwidth}
  \includegraphics[trim=0cm 0cm 0cm 0cm, clip, width=\linewidth]{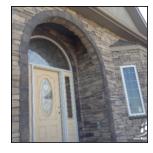}
  \caption*{stone wall}
\end{subfigure}\ \ 
\begin{subfigure}{0.19\textwidth}
  \includegraphics[trim=0cm 0cm 0cm 0cm, clip, width=\linewidth]{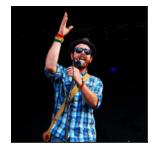}
  \caption*{sunglasses}
\end{subfigure}\ \ 
\begin{subfigure}{0.19\textwidth}
  \includegraphics[trim=0cm 0cm 0cm 0cm, clip, width=\linewidth]{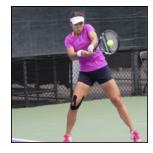}
  \caption*{tennis ball}
\end{subfigure}\ \ 
\begin{subfigure}{0.19\textwidth}
  \includegraphics[trim=0cm 0cm 0cm 0cm, clip, width=\linewidth]{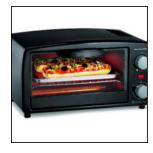}
  \caption*{toaster}
\end{subfigure}\ \ 
\begin{subfigure}{0.19\textwidth}
  \includegraphics[trim=0cm 0cm 0cm 0cm, clip, width=\linewidth]{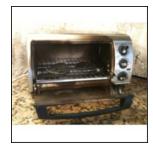}
  \caption*{toaster}
\end{subfigure} 
\caption{Examples of images (from the heldout-debug set) on which $20$ highly accurate ImageNet trained models predict incorrectly (ground truth label below each image). However, models trained using our framework make correct predictions on all of them.}
\label{fig:appendix_images_90_106}
\end{figure*}

Figure \ref{fig:appendix_images_90_106} shows images several from the heldout-debug set on which we obtain correct predictions.

\clearpage

\section{Comparing between images discovered using weak labels and no labels}\label{sec:appendix_compare_weaklabel_nolabel}

\begin{figure}[ht!]
        \centering
        \begin{minipage}{0.16\textwidth}
        \hspace*{0.4cm}\begin{subfigure}{0.9\linewidth}
        \includegraphics[trim=0cm 0cm 0cm 0cm, clip, width=\linewidth]{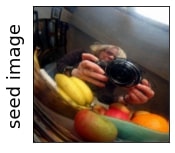}
        \caption*{banana}
        \end{subfigure}
        \end{minipage}
        \hfill
        \begin{minipage}{0.80\textwidth}
        \begin{subfigure}{\linewidth}
        \includegraphics[trim=0cm 0cm 0cm 0cm, clip, width=\linewidth]{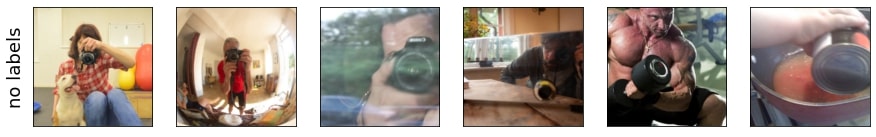}
        \end{subfigure}
        \begin{subfigure}{\linewidth}
        \vspace{-0.2cm}
        \includegraphics[trim=0cm 0cm 0cm 0cm, clip, width=\linewidth]{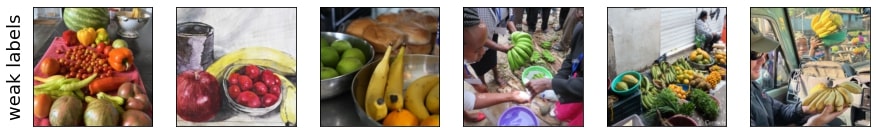}
        \end{subfigure}
        \end{minipage}
        \vspace{-0.1cm}
    \caption{left: image with label \textbf{banana}, right: similar images obtained using the two different methods}
    \label{fig:appendix_comparison_0}
\end{figure}

\begin{figure}[ht!]
        \centering
        \captionsetup{type=figure}
        \begin{minipage}{0.16\textwidth}
        \hspace*{0.4cm}\begin{subfigure}{0.9\linewidth}
        \includegraphics[trim=0cm 0cm 0cm 0cm, clip, width=\linewidth]{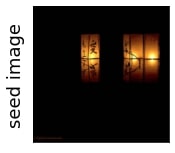}
        \caption*{candle}
        \end{subfigure}
        \end{minipage}
        \hfill
        \begin{minipage}{0.80\textwidth}
        \begin{subfigure}{\linewidth}
        \includegraphics[trim=0cm 0cm 0cm 0cm, clip, width=\linewidth]{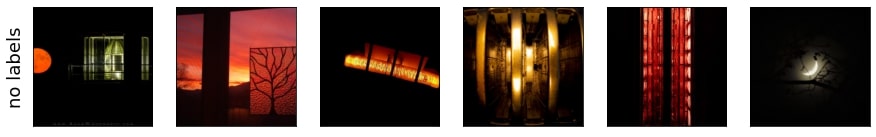}
        \end{subfigure}
        \begin{subfigure}{\linewidth}
        \vspace{-0.2cm}
        \includegraphics[trim=0cm 0cm 0cm 0cm, clip, width=\linewidth]{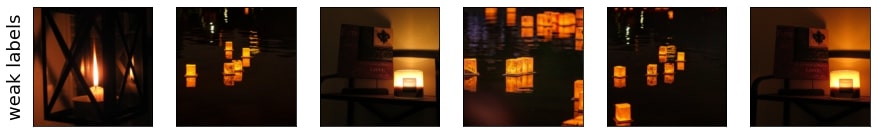}
        \end{subfigure}
        \end{minipage}
        \vspace{-0.1cm}
    \caption{left: image with label \textbf{candle}, right: similar images obtained using the two different methods}
    \label{fig:appendix_comparison_1}
\end{figure}

\begin{figure}
        \centering
        \captionsetup{type=figure}
        \begin{minipage}{0.16\textwidth}
        \hspace*{0.4cm}\begin{subfigure}{0.9\linewidth}
        \includegraphics[trim=0cm 0cm 0cm 0cm, clip, width=\linewidth]{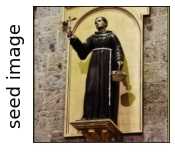}
        \caption*{church}
        \end{subfigure}
        \end{minipage}
        \hfill
        \begin{minipage}{0.80\textwidth}
        \begin{subfigure}{\linewidth}
        \includegraphics[trim=0cm 0cm 0cm 0cm, clip, width=\linewidth]{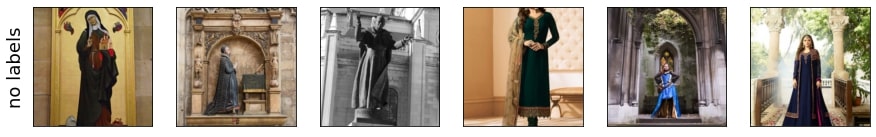}
        \end{subfigure}
        \begin{subfigure}{\linewidth}
        \vspace{-0.2cm}
        \includegraphics[trim=0cm 0cm 0cm 0cm, clip, width=\linewidth]{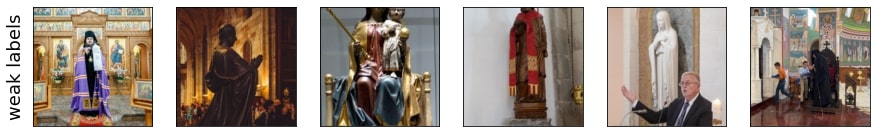}
        \end{subfigure}
        \end{minipage}
        \vspace{-0.1cm}
    \caption{left: image with label \textbf{church}, right: similar images obtained using the two different methods}
    \label{fig:appendix_comparison_2}
\end{figure}

\begin{figure}
        \centering
        \captionsetup{type=figure}
        \begin{minipage}{0.16\textwidth}
        \hspace*{0.4cm}\begin{subfigure}{0.9\linewidth}
        \includegraphics[trim=0cm 0cm 0cm 0cm, clip, width=\linewidth]{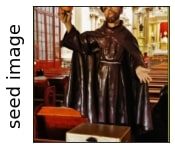}
        \caption*{church}
        \end{subfigure}
        \end{minipage}
        \hfill
        \begin{minipage}{0.80\textwidth}
        \begin{subfigure}{\linewidth}
        \includegraphics[trim=0cm 0cm 0cm 0cm, clip, width=\linewidth]{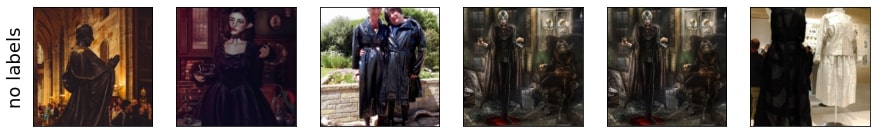}
        \end{subfigure}
        \begin{subfigure}{\linewidth}
        \vspace{-0.2cm}
        \includegraphics[trim=0cm 0cm 0cm 0cm, clip, width=\linewidth]{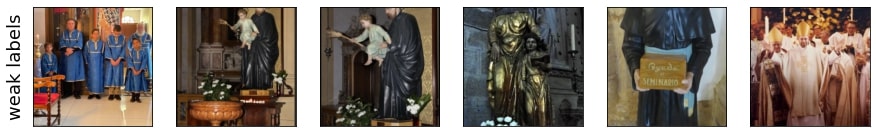}
        \end{subfigure}
        \end{minipage}
        \vspace{-0.1cm}
    \caption{left: image with label \textbf{church}, right: similar images obtained using the two different methods}
    \label{fig:appendix_comparison_3}
\end{figure}

\begin{figure}
        \centering
        \captionsetup{type=figure}
        \begin{minipage}{0.16\textwidth}
        \hspace*{0.4cm}\begin{subfigure}{0.9\linewidth}
        \includegraphics[trim=0cm 0cm 0cm 0cm, clip, width=\linewidth]{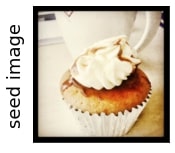}
        \caption*{coffee mug}
        \end{subfigure}
        \end{minipage}
        \hfill
        \begin{minipage}{0.80\textwidth}
        \begin{subfigure}{\linewidth}
        \includegraphics[trim=0cm 0cm 0cm 0cm, clip, width=\linewidth]{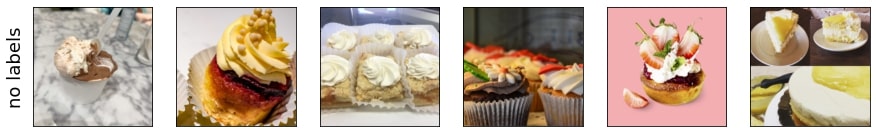}
        \end{subfigure}
        \begin{subfigure}{\linewidth}
        \vspace{-0.2cm}
        \includegraphics[trim=0cm 0cm 0cm 0cm, clip, width=\linewidth]{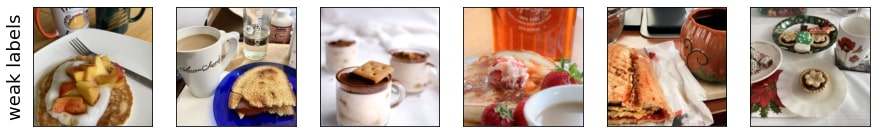}
        \end{subfigure}
        \end{minipage}
        \vspace{-0.1cm}
    \caption{left: image with label \textbf{coffee mug}, right: similar images obtained using the two different methods}
    \label{fig:appendix_comparison_4}
\end{figure}

\begin{figure}
        \centering
        \captionsetup{type=figure}
        \begin{minipage}{0.16\textwidth}
        \hspace*{0.4cm}\begin{subfigure}{0.9\linewidth}
        \includegraphics[trim=0cm 0cm 0cm 0cm, clip, width=\linewidth]{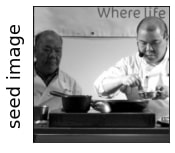}
        \caption*{frypan}
        \end{subfigure}
        \end{minipage}
        \hfill
        \begin{minipage}{0.80\textwidth}
        \begin{subfigure}{\linewidth}
        \includegraphics[trim=0cm 0cm 0cm 0cm, clip, width=\linewidth]{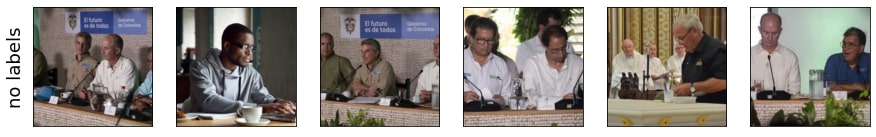}
        \end{subfigure}
        \begin{subfigure}{\linewidth}
        \vspace{-0.2cm}
        \includegraphics[trim=0cm 0cm 0cm 0cm, clip, width=\linewidth]{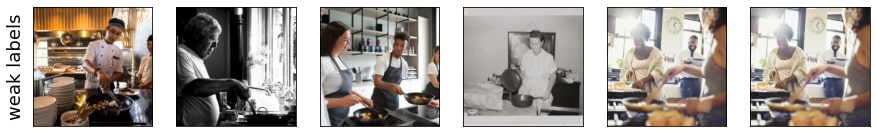}
        \end{subfigure}
        \end{minipage}
        \vspace{-0.1cm}
    \caption{left: image with label \textbf{frypan}, right: similar images obtained using the two different methods}
    \label{fig:appendix_comparison_5}
\end{figure}

\begin{figure}
        \centering
        \captionsetup{type=figure}
        \begin{minipage}{0.16\textwidth}
        \hspace*{0.4cm}\begin{subfigure}{0.9\linewidth}
        \includegraphics[trim=0cm 0cm 0cm 0cm, clip, width=\linewidth]{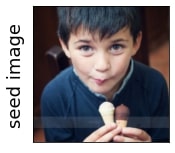}
        \caption*{ice cream}
        \end{subfigure}
        \end{minipage}
        \hfill
        \begin{minipage}{0.80\textwidth}
        \begin{subfigure}{\linewidth}
        \includegraphics[trim=0cm 0cm 0cm 0cm, clip, width=\linewidth]{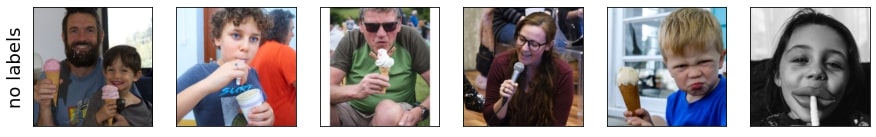}
        \end{subfigure}
        \begin{subfigure}{\linewidth}
        \vspace{-0.2cm}
        \includegraphics[trim=0cm 0cm 0cm 0cm, clip, width=\linewidth]{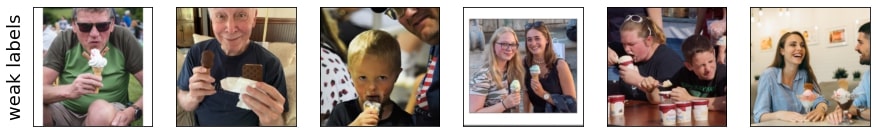}
        \end{subfigure}
        \end{minipage}
        \vspace{-0.1cm}
    \caption{left: image with label \textbf{ice cream}, right: similar images obtained using the two different methods}
    \label{fig:appendix_comparison_6}
\end{figure}

\begin{figure}
        \centering
        \captionsetup{type=figure}
        \begin{minipage}{0.16\textwidth}
        \hspace*{0.4cm}\begin{subfigure}{0.9\linewidth}
        \includegraphics[trim=0cm 0cm 0cm 0cm, clip, width=\linewidth]{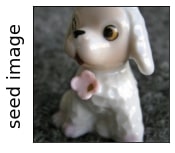}
        \caption*{toy poodle}
        \end{subfigure}
        \end{minipage}
        \hfill
        \begin{minipage}{0.80\textwidth}
        \begin{subfigure}{\linewidth}
        \includegraphics[trim=0cm 0cm 0cm 0cm, clip, width=\linewidth]{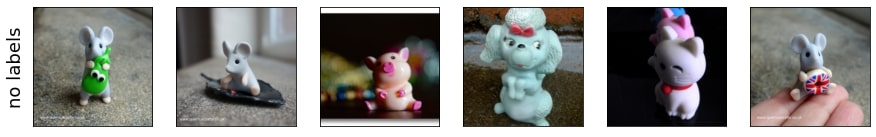}
        \end{subfigure}
        \begin{subfigure}{\linewidth}
        \vspace{-0.2cm}
        \includegraphics[trim=0cm 0cm 0cm 0cm, clip, width=\linewidth]{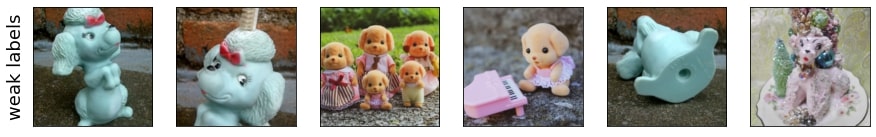}
        \end{subfigure}
        \end{minipage}
        \vspace{-0.1cm}
    \caption{left: image with label \textbf{toy poodle}, right: similar images obtained using the two different methods}
    \label{fig:appendix_comparison_7}
\end{figure}

\begin{figure}
        \centering
        \captionsetup{type=figure}
        \begin{minipage}{0.16\textwidth}
        \hspace*{0.4cm}\begin{subfigure}{0.9\linewidth}
        \includegraphics[trim=0cm 0cm 0cm 0cm, clip, width=\linewidth]{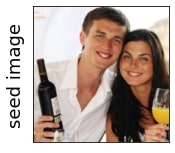}
        \caption*{wine bottle}
        \end{subfigure}
        \end{minipage}
        \hfill
        \begin{minipage}{0.80\textwidth}
        \begin{subfigure}{\linewidth}
        \includegraphics[trim=0cm 0cm 0cm 0cm, clip, width=\linewidth]{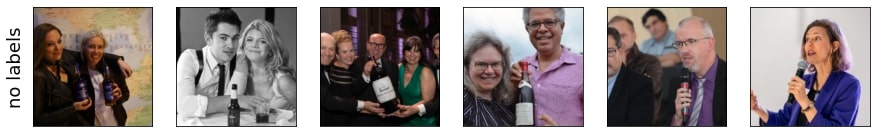}
        \end{subfigure}
        \begin{subfigure}{\linewidth}
        \vspace{-0.2cm}
        \includegraphics[trim=0cm 0cm 0cm 0cm, clip, width=\linewidth]{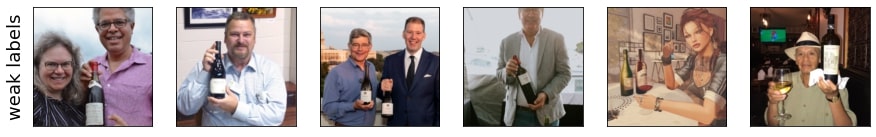}
        \end{subfigure}
        \end{minipage}
        \vspace{-0.1cm}
    \caption{left: image with label \textbf{wine bottle}, right: similar images obtained using the two different methods}
    \label{fig:appendix_comparison_8}
\end{figure}

\end{document}

%% file: commands.tex
\usepackage{dsfont}
\usepackage{enumitem}

\usepackage{soul}
\usepackage{subcaption}
\usepackage{multirow}
\usepackage{booktabs}
\usepackage{csquotes}
\usepackage{graphicx}
\usepackage{mathtools}

\newcommand{\dist}{\mathcal{E}}

\newcommand{\set}{\mathcal{A}}

\DeclareMathOperator*{\argmax}{arg\,max}

\newcommand{\cX}{\mathcal{X}}
\newcommand{\cY}{\mathcal{Y}}
\newcommand{\bx}{\mathbf{x}}

\newcommand{\bz}{\mathbf{z}}

\newcommand{\allclasses}{\mathcal{Y}}
\newcommand{\errorclasses}{\mathcal{T}}

\newcommand{\query}{\mathcal{Q}}
\newcommand{\union}{\mathcal{U}}
\newcommand{\web}{\mathcal{W}}

\newcommand{\val}{\mathcal{V}}

\newcommand{\near}{\mathcal{N}}
\newcommand{\flickr}{\mathcal{F}}
\newcommand{\imagenet}{\mathcal{I}}

\usepackage{enumitem}

